\documentclass[runningheads]{llncs}
\usepackage{graphicx}

\usepackage{tikz}
\usepackage{comment}
\usepackage{amsmath,amssymb} 
\usepackage{color}
\usepackage{orcidlink}

\usepackage{caption, subcaption}
\usepackage{algorithm}
\usepackage{algpseudocode}

\usepackage[accsupp]{axessibility}  

\usepackage{amsmath,amsfonts,bm}

\def\eqref#1{equation~\ref{#1}}

\def\1{\bm{1}}

\newcommand{\test}{\mathcal{D_{\mathrm{test}}}}

\def\rvx{{\mathbf{x}}}

\def\rvz{{\mathbf{z}}}
\def\rvzeta{{\mathbf{\zeta}}}

\DeclareMathAlphabet{\mathsfit}{\encodingdefault}{\sfdefault}{m}{sl}
\SetMathAlphabet{\mathsfit}{bold}{\encodingdefault}{\sfdefault}{bx}{n}

\newcommand{\E}{\mathbb{E}}

\newcommand{\R}{\mathbb{R}}

\newcommand{\KL}{D_{\mathrm{KL}}}

\newcommand{\Cov}{\mathrm{Cov}}

\begin{document}
\pagestyle{headings}
\mainmatter
\def\ECCVSubNumber{6895}  

\newcommand{\etal}{et al.}
\newcommand{\rulesep}{\unskip\ \vrule\ }

\title{NashAE: Disentangling Representations through Adversarial Covariance Minimization} 

\titlerunning{NashAE}
%
\author{Eric Yeats\inst{1}\orcidlink{0000-0002-5470-3380} \and
Frank Liu\inst{2}\orcidlink{0000-0001-6615-0739} \and
David Womble\inst{2}\orcidlink{0000-0003-1741-8983} \and
Hai Li\inst{1}\orcidlink{0000-0003-3228-6544}}
\authorrunning{E. Yeats et al.}
%
\institute{
Duke University, Durham NC 27708 \\
\email{\{eric.yeats, hai.li\}@duke.edu}
\and
Oak Ridge National Laboratory, Oak Ridge TN 37830 \\
\email{\{liufy, womblede\}@ornl.gov}
}

\maketitle

\begin{abstract}
We present a self-supervised method to disentangle factors of variation in high-dimensional data that does not rely on prior knowledge of the underlying variation profile (e.g., no assumptions on the number or distribution of the individual latent variables to be extracted). In this method which we call NashAE, high-dimensional feature disentanglement is accomplished in the low-dimensional latent space of a standard autoencoder (AE) by promoting the discrepancy between each encoding element and information of the element recovered from all other encoding elements. Disentanglement is promoted efficiently by framing this as a minmax game between the AE and an ensemble of regression networks which each provide an estimate of an element conditioned on an observation of all other elements. We quantitatively compare our approach with leading disentanglement methods using existing disentanglement metrics. Furthermore, we show that NashAE has increased reliability and increased capacity to capture salient data characteristics in the learned latent representation.
\keywords{representation learning, autoencoder, adversarial, minmax game}
\end{abstract}

\section{Introduction}

\begin{figure}[ht]
    \centering
    \includegraphics[width=0.5\columnwidth]{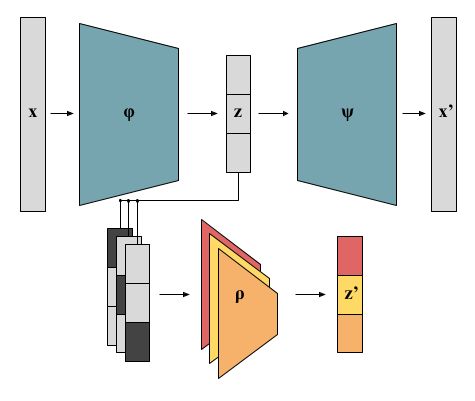}
    \caption{Depiction of the proposed disentanglement method, NashAE, for a latent space dimensionality of $m=3$. An autoencoder (AE), composed of the encoder $\phi$ and decoder $\psi$, compresses a high dimensional input $\rvx \in \R^n$ to a lower dimensional latent vector $\rvz \in \R^m$, and decompresses $\rvz$ to approximate $\rvx$ as $\rvx'$. An ensemble of $m$ independently trained regression networks $\rho$ takes $m$ duplicates of $\rvz$ which each have an element removed, and each independent regression network tries to predict the value of its missing element using knowledge of the other elements. Disentanglement is achieved through equilibrium in an adversarial game in which $\phi$ minimizes the element-wise covariance between the true latent vector $\rvz$ and concatenated predictions vector $\rvz'$}
\end{figure}

Deep neural networks (DNNs) have proven to be extremely high-performing in the realms of computer vision \cite{he2016deep}, natural language processing \cite{torfi2020natural}, autonomous control \cite{li2019reinforcement}, and deep generative models \cite{goodfellow2014generative,kingma2013auto}, among others. The huge successes of DNNs have made them almost ubiquitous as an engineering tool, and it is very common for them to appear in many new applications. However, as we rush to deploy DNNs in the real world, we have also exposed many of their shortcomings. 

One such shortcoming is that DNNs are extremely sensitive to minute perturbations in their inputs \cite{goodfellow2014explaining,szegedy2013intriguing} or weights \cite{he2020defending}, causing otherwise high-performing models to suddenly be consistently incorrect. Additionally, DNNs trained on image classification tasks are observed to predict labels confidently even when the image shares no relationship with their in-distribution label space \cite{hendrycks2016baseline,liang2017enhancing}. Furthermore, DNNs are known to perpetuate biases in their training data through their predictions, exacerbating salient issues such as racial and social inequity, and gender inequality \cite{alvi2018turning}. While this small subset of examples may appear to be unrelated, they are all linked by a pervasive issue of DNNs: their lack of interpretability \cite{ross2018improving,rudin2019stop}. The fact that DNNs are treated as black boxes, where engineers lack a clear explanation or reasoning for why or how DNNs make decisions, makes the root cause of DNNs' shortcomings difficult to diagnose.

A promising remedy to this overarching issue is to clarify learning representations through feature disentanglement: the process of learning unique data representations that are each only sensitive to independent factors of the underlying data distribution. It follows that a disentangled representation is an inherently interpretable representation, where each disentangled unit has a consistent, unique, and independent interpretation of the data across its domain.

Several works have pioneered the field of feature disentanglement, moving from supervised \cite{kulkarni2015deep} to unsupervised approaches \cite{burgess2018understanding,chen2016infogan,higgins2016beta,hu2018disentangling}, which we focus on in this work. Chen \etal~\cite{chen2016infogan} present InfoGAN, an extension of the GAN framework \cite{goodfellow2014generative} that enables better control of generated images using special noise variables. Higgins \etal~\cite{higgins2016beta} introduce $\beta$-VAE, a generalization of the VAE framework \cite{kingma2013auto} that allows the VAE to extract more statistically independent, disentangled representations.

While these highly successful methods are considered to be unsupervised, they still have a considerable amount of prior knowledge built into their operation. InfoGAN requires prior knowledge of the number and form of disentangled factors to extract, and $\beta$-VAE encounters bottleneck capacity issues and inconsistent results with seemingly innocuous changes to hyperparameters, requiring finetuning with some supervision \cite{burgess2018understanding,higgins2016beta}.

We propose a new method, NashAE, to promote a sparse and disentangled latent space in the standard AE that does not make assumptions on the number or distribution of underlying data generating factors. The core intuition behind the approach is to reduce the redundant information between latent encoding elements, regardless of their distribution. To accomplish this, this work presents a new technique to reduce the information between encoded continuous and/or discrete random variables using just access to samples drawn from the unknown underlying distributions. We empirically demonstrate that the method can reliably extract high-quality disentangled representations according to the metric presented in \cite{higgins2016beta}, and that the method has a higher latent feature capacity with respect to salient data characteristics.

The paper makes the following contributions:
\begin{itemize}
    \item We develop a method to quantify the relationship between random variables with unknown distribution (arbitrary continuous or discrete/categorical), and show that it can be used to promote statistical independence among latent variables in an AE. 
    \item We provide qualitative and quantitative evidence that NashAE reliably extracts a set of disentangled continuous and/or discrete factors of variation in a variety of scenarios, and we demonstrate the method's improved latent feature capacity with regard to salient data characteristics.
    \item We release the Beamsynthesis disentanglement dataset, a collection of time-series data based on particle physics studies and their associated data generating factor ground truth.
\end{itemize}

The code for all experiments and the Beamsynthesis dataset can be found at: \url{https://github.com/ericyeats/nashae-beamsynthesis}.

\section{Related Work}

\textbf{Autoencoders.} Much of this work derives from autoencoders (AE), which consist of an encoder function followed by a decoder function. The encoder transforms high-dimensional input $\rvx \sim X$ into a low-dimensional latent representation $\rvz$, and the decoder transforms $\rvz$ to a reconstruction of the high-dimensional input $\rvx'$. AE have numerous applications in the form of unsupervised feature learning, denoising, and feature disentanglement \cite{hu2018disentangling,lu2013speech,rifai2011higher}. Variational autoencoders (VAEs) \cite{kingma2013auto} take AEs further by using them to parameterize probability distributions for $X$. VAEs are trained by maximizing a lower bound of the likelihood of $X$, a process which involves conforming the encoded latent space $\rvz \sim Z$ with a prior distribution $P$. Adversarial AEs \cite{makhzani2015adversarial}, like VAEs, match encoded distributions to a prior distribution, but do so through an adversarial procedure inspired by Generative Adversarial Networks (GANs) \cite{goodfellow2014generative}.

\textbf{Unsupervised Disentanglement Methods.} One of the most successful approaches to feature disentanglement is $\beta$-VAE \cite{higgins2016beta}, which builds on the VAE framework. $\beta$-VAE adjusts the VAE training framework by modulating the relative strength of the $\KL(Z||P)$ term with hyperparameter $\beta$, effectively limiting the capacity of the VAE and encouraging disentanglement as $\beta$ becomes larger. Higgins \etal~\cite{higgins2016beta} note a positive correlation between the size of the VAE latent dimension and the optimal $\beta$ hyperparameter to do so, requiring some hyperparameter search and limited supervision. Another important contribution of Higgins \etal~\cite{higgins2016beta} is a metric for quantifying disentanglement which depends on the accuracy of a linear classifier in determining which data generating factor is held constant over a pair of data batches. 

Multiple works have augmented $\beta$-VAE with loss functions that isolate the Total Correlation (TC) component of $\KL(Z||P)$, further boosting quantitative disentanglement performance in certain scenarios \cite{chen2018isolating,kim2018disentangling}. Another VAE-based work proposed by Kumar \etal~\cite{kumar2017variational} directly minimizes the covariance of the encoded representation. However, simple covariance of the latent elements fails to capture more complex, nonlinear relationships between the elements. Our work employs regression neural networks to capture complex dependencies.

Chen \etal~\cite{chen2016infogan} present InfoGAN, which builds on the GAN framework \cite{goodfellow2014generative}. InfoGAN augments the base GAN training procedure with a special set of independent noise inputs. A tractable lower bound on MI is maximized between the special noise inputs and output of the generator, leading to the special noise inputs resembling data generating factors. While the method claims to be unsupervised, choosing its special noise inputs requires prior knowledge of the number and nature (e.g. distribution) of factors to extract.

\textbf{Limitations of Unsupervised Disentanglement.} Locatello \etal~\cite{locatello2019challenging} demonstrate that unsupervised disentanglement learning is fundamentally impossible without incorporating inductive biases on both models and data \cite{locatello2019challenging}. However, they assert that given the right inductive biases, the prospect of unsupervised disentanglement learning is not so bleak. We incorporate several inductive biases in our method to achieve unsupervised disentanglement. First, our approach assumes that disentangled learning representations are characterised by being statistically independent. Second, we posit that breaking up the latent factorization problem into multiple parts by individual masking and adversarial covariance minimization helps boost disentanglement reliability. In terms of models and data, we employ the network architectures and data preparation suggested by previous works in unsupervised disentanglement. Under such conditions, NashAE has demonstrated superior reliability in retrieving disentangled representations.

\section{NashAE Methodology}\label{sec:methodology}

Our approach starts with a purely deterministic encoder $\phi$, which takes input observations $\rvx \sim X$ and creates a latent representation $\rvz = \phi(\rvx)$. Where $\rvx \in \R^n$, $\rvz \in \R^m$, and typically $n \gg m$. Furthermore, $\phi$ employs a sigmoid activation function $\sigma$ at its output to produce $\rvz$ such that $\rvz = \sigma(\rvzeta)$ and $\rvz \in [0, 1]^m$, where $\rvzeta$ is the output of $\phi$ before it is passed through the sigmoid non-linearity. A deterministic decoder $\psi$ maps the latent representation $\rvz$ back to the observation domain $\rvx' = \psi(\rvz)$. To achieve disentanglement, the AE is trained with two complementary objectives: (1) reconstructing the observations, and (2) maximizing the discrepancy between each latent variable and predicted values of the variable using information of all other variables. The intuitions behind each are the following. First, reconstruction of the input observations $\rvx$ is standard of AEs and ensures that they learn features relevant to the distribution $X$. Second, promoting discrepancy between $i$-th latent element and its prediction (conditioned on all other $j \neq i$ elements) reduces the information between latent element $i$ and all other elements $j \neq i$.

For the reconstruction objective, the goal is to minimize the mean squared error:

\begin{equation}\label{eqn:mse_ae}
    \mathcal{L}_R = \frac{1}{2 n}\E_{\rvx \sim X}\big|\big|\rvx' - \rvx\big|\big|^2_2 .
\end{equation}

Reconstructing the input observation $\rvx$ ensures that the features of the latent space are relevant to the underlying data distribution $X$. The following subsection describes the adversarial game loss objectives, which settle on an equilibrium and inspire the name of the proposed disentanglement method, NashAE.

\subsection*{Adversarial Covariance Minimization}

In general, it is difficult to compute the information between latent variables when one only has access to samples of observations $\rvz \sim Z$. Since the underlying distribution $Z$ is unknown, standard methods of computing the information directly are not possible. To overcome this challenge, we propose to reduce the information between latent variables indirectly using an ensemble of regression networks which attempt to capture the relationships between latent variables. The process is computationally efficient; it uses simple measures of linear statistical independence and an adversarial game.

Consider an ensemble of $m$ independent regression networks $\rho$, where the output of the $i$-th network $\rho_i$ corresponds to a missing $i$-th latent element. The objective of each $\rho_i$ is to minimize the expected squared error:

\begin{equation}\label{eqn:mse_pred}
     \mathcal{L}_{\rho_i} = \frac{1}{2}\E_{\rvx \sim X}\big(\rho(\overline{\rvz_i})_i - \rvz_i\big)^2 ,
\end{equation}

where $\rvz_i$ is the $i$-th true latent element, and $\overline{\rvz_i}$ is the latent \textit{vector} with the $i$-th latent element masked with $0$ (i.e., all elements of the latent vector are present except $\rvz_i$).

We call $\rho$ the predictors, since they are each optimized to predict one missing value of $\rvz$ given knowledge of all other $\rvz$. If all their individual predictions are concatenated together, they form $\rvz'$ such that each $\rvz'_i = \rho(\overline{\rvz_i})_i$.

For the disentanglement objective, we want to choose encodings $\rvz \sim \phi(X)$ that make it difficult to recover information of one element from all others. This leads to a natural minmax formulation for the AE and predictors:

\begin{equation}\label{eqn:minmax}
    \min_{\phi,\psi} \max_{\rho} \frac{1}{2} \E_{\rvx \sim X} \left[\frac{1}{n}\big|\big|\rvx' - \rvx\big|\big|^2_2 - \big|\big|\rvz' - \rvz\big|\big|^2_2 \right] .
\end{equation}

In general, each predictor attempts to use information of $\overline{\rvz_i}$ to establish a one-to-one linear relationship between $\rvz'$ and $\rvz$. Hence we propose to use covariance between $\rvz'$ and $\rvz$ across a batch of examples to capture the degree to which they are related. In practice, we find that training the AE to minimize the summed covariance objective between each of the $\rvz_i'$ and $\rvz_i$ random variable pairs,

\begin{equation}\label{eqn:cov_ae}
    \mathcal{L}_A = \sum_{i=1}^m\Cov(\rvz'_i, \rvz_i) ,
\end{equation}
is more stable than maximizing $\frac{1}{2}\E_{\rvx \sim X}||\rvz' - \rvz||^2_2$ and leads to more reliable disentanglement outcomes. Hence, in all the following experiments we train the AE to minimize this summed covariance measure, $\mathcal{L}_A$. Furthermore, one can show that the fixed points of the minmax objective (\ref{eqn:minmax}) are the same as those of training $\phi$ to minimize $\mathcal{L}_R + \mathcal{L}_A$ for disentangled representations (see supplementary material).

In the adversarial loss $\mathcal{L}_A$, the optimization objective of the encoder $\phi$ is to adjust its latent representations to minimize the covariance between each  $\rvz'_i$ and $\rvz_i$. Using minibatch stochastic gradient descent (SGD), the encoder $\phi$ can use gradient passed through the predictors $\rho$ to learn exactly how to adjust its latent representations to minimize the adversarial loss. Assuming that $\rho$ can learn faster than $\phi$, each $i$-th covariance term will reach zero when $\E[\rvz_i|\rvz_j, \forall j \neq i] = \E[\rvz_i]$ everywhere.

In the following experiments, we weight the sum of $\mathcal{L}_R$ and $\mathcal{L}_A$ with the hyperparameter $\lambda \in [0, 1)$ in order to establish a normalized balance between the reconstruction and adversarial objectives:
\begin{equation}\label{eqn:mse_cov_ae}
    \mathcal{L}_{R,A}(\lambda) = (1 - \lambda) \mathcal{L}_R + \lambda \mathcal{L}_A .
\end{equation}

Intuitively, higher values of $\lambda$ result lower covariance between elements of $\rvz$ and $\rvz'$, and eventually the equilibrium covariance settles to zero. 
In the special case where all data generating factors are independent, the AE can theoretically achieve $\mathcal{L}_{R,A}=0$.

\subsection*{Proposed Disentanglement Metric: TAD}

In the following section, we find that NashAE and $\beta$-VAE could achieve equally high scores using the $\beta$-VAE metric. However, the $\beta$-VAE metric fails to capture a key aspect of a truly disentangled latent representation: change in one independent data generating factor should correspond to change in just one disentangled latent feature. This is not captured in the $\beta$-VAE disentanglement metric since the score can benefit from \textit{spreading} the information of one data generating factor over multiple latent features. For example, duplicate latent representations of the same unique data generating factor can only increase the score of the $\beta$-VAE metric.

Furthermore, a disentanglement metric should quantify the degree to which its set of independent latent axes aligns with the independent data generating factor ground truth axes. In essence, a unique latent feature should be a confident predictor of a unique data generating factor, and all other latents should be orthogonal to the same data generating factor. Intuitively, the greater number of latent axes that align uniquely with the data generating factors and the more confident the latents are as predictors of the factors, the higher the metric score should be. 

For these reasons, we design a disentanglement metric for datasets with binary attribute ground truth labels called Total AUROC Difference (TAD). For a large number $l$ of examples which we collect a batch of latent representations $z$ of the shape $(l, m)$, we perform the following to calculate the TAD:
\begin{enumerate}
    \item For each independent ground truth attribute, calculate the AUROC score of each latent variable in detecting the attribute.
    \item For each independent ground truth attribute, find the maximum latent AUROC score $a_{1,i}$ and the next-largest latent AUROC score $a_{2,i}$, where $i$ is the index of the independent ground truth attribute under consideration.
    \item Take $\sum_i a_{1,i}-a_{2,i}$ as the TAD score, where $i$ indexes over the independent ground truth attributes.
\end{enumerate}

The TAD metric captures important aspects of a disentangled latent representation. First, each AUROC difference $a_{1,i}-a_{2,i}$ captures the degree to which a unique attribute is detected by a unique latent representation. Second, summing the AUROC difference scores for each independent ground truth attribute quantifies the degree to which the latent axes confidently replicate the ground truth axes. See the supplementary material for more details on how TAD is calculated and for a discussion relating it with other work. 

\section{Experiments}

The following section contains a mix of qualitative and quantitative results for four unsupervised disentanglement algorithms: NashAE (this work), $\beta$-VAE \cite{higgins2016beta}, FactorVAE \cite{kim2018disentangling}, and $\beta$-TCVAE \cite{chen2018isolating}. The results are collected for disentanglement tasks on three datasets: Beamsynthesis, dSprites \cite{higgins2016beta}, and CelebA \cite{liu2015faceattributes}. Please refer to the supplementary material for details on algorithm hyperparameters, network architectures, and data normalization for the different experiments.

\textbf{Beamsynthesis} is a simple dataset of 360 time-series data of current waveforms constructed from simulations of the LINAC (linear particle accelerator) portion of high-energy particle accelerators. The dataset contains two ground truth data generating factors: a categorical random variable representing the \textit{frequency} of the particle waveform which can take on one of the three values ($10, 15, 20$) and a continuous random variable constructed from a uniform sweep of 120 waveform \textit{duty cycle} values $\in [0.2, 0.8)$. The Cartesian product of the two data generating factors forms the set of observations. The challenge in disentangling this dataset arises from the fact that both the \textit{frequency} and \textit{duty cycle} of a waveform affect the length of the "on" period of each wave. We visualize the complete latent space of different algorithms and evaluate the reliability of the algorithms in extracting the correct number of ground truth data generating factors using this dataset.

\textbf{dSprites} is a disentanglement dataset released by the authors of $\beta$-VAE - it is comprised of a set of $737,280$ images of white 2D shapes on a black background. The Cartesian product of the type of shape (categorical: square, ellipse, heart), scale (continuous: 6 values), orientation (continuous: 40 values), x-position (continuous: 32 values), and y-position (continuous: 32 values) forms the independent ground truth of the dataset. We measure the $\beta$-VAE disentanglement metric score for different algorithms using this dataset.

\textbf{CelebA} is a dataset comprised of $202,599$ images of the faces of $10,177$ different celebrities. Associated with each image are 40 different binary attribute labels such as \textit{bangs}, \textit{blond hair}, \textit{black hair}, \textit{chubby}, \textit{male}, and \textit{eyeglasses}. We measure the TAD score of different algorithms using this dataset.

\subsection*{Empirical Fixed Point Results}

In section \ref{sec:methodology}, we indicate that higher values of $\lambda \in (0, 1)$ should result in a statistically independent NashAE latent space, and that redundant latent elements will not be learned. This is supported by observations of the fixed point of the optimization process for all experiments with nonzero $\lambda$: as $\lambda$ is increased, the number of dead latent representations increases, and the average $R^2$ correlation statistic between latent representations and their predictions decreases.

\begin{figure*}[ht]
\centering
\begin{subfigure}[t]{0.31\textwidth}
    \centering
    \includegraphics[width=\textwidth]{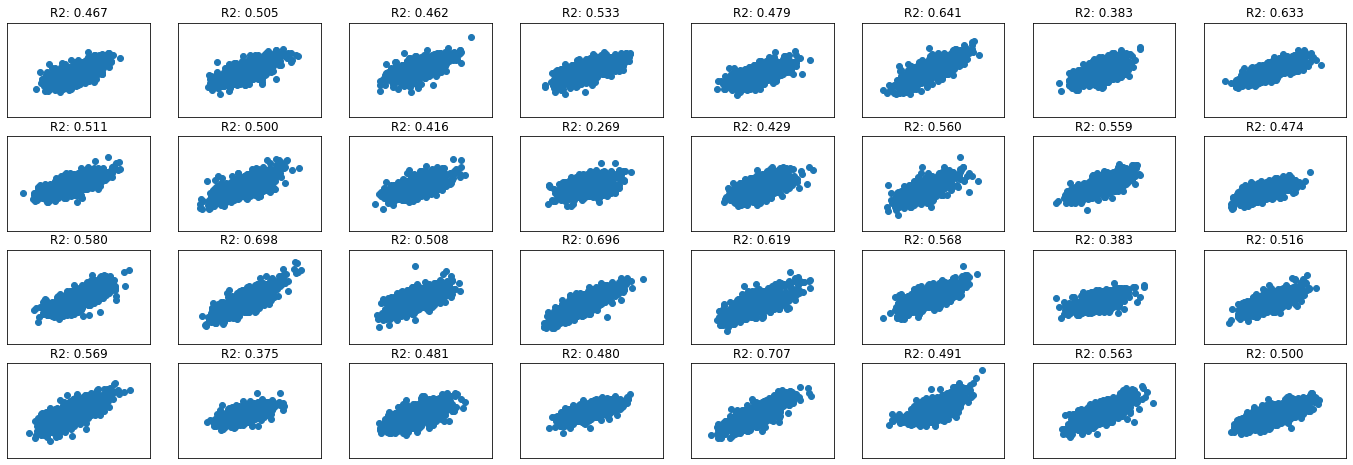}
    \caption{$\lambda=0.0$. The average $R^2$ statistic is 0.52, and 32 latent representations are learned.}
\end{subfigure}
\rulesep
\begin{subfigure}[t]{0.31\textwidth}
    \centering
    \includegraphics[width=\textwidth]{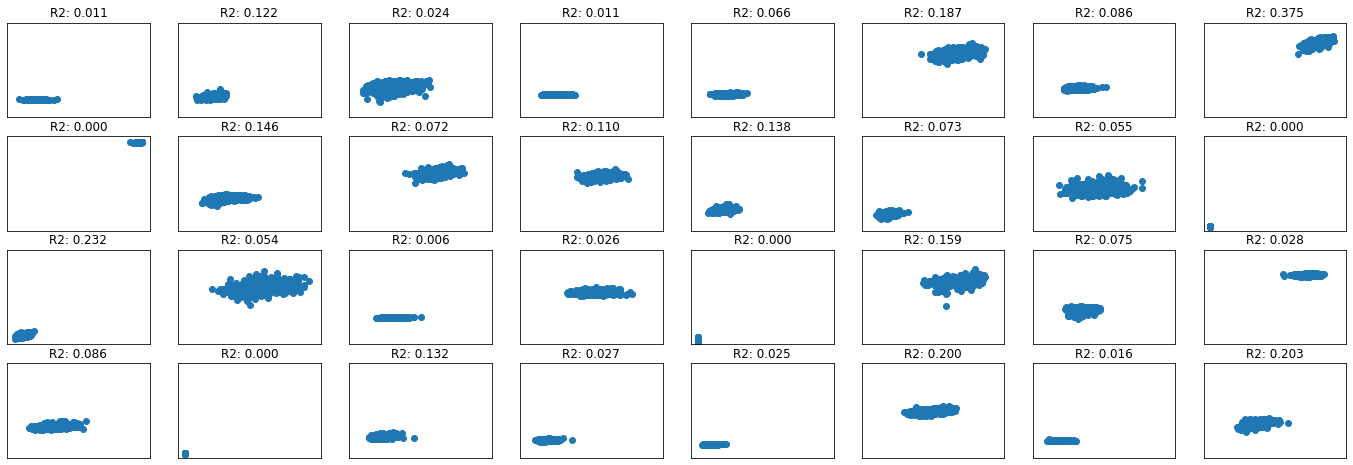}
    \caption{$\lambda=0.1$. The average $R^2$ statistic is 0.09, and 28 latent representations are learned.}
\end{subfigure}
\rulesep
\begin{subfigure}[t]{0.31\textwidth}
    \centering
    \includegraphics[width=\textwidth]{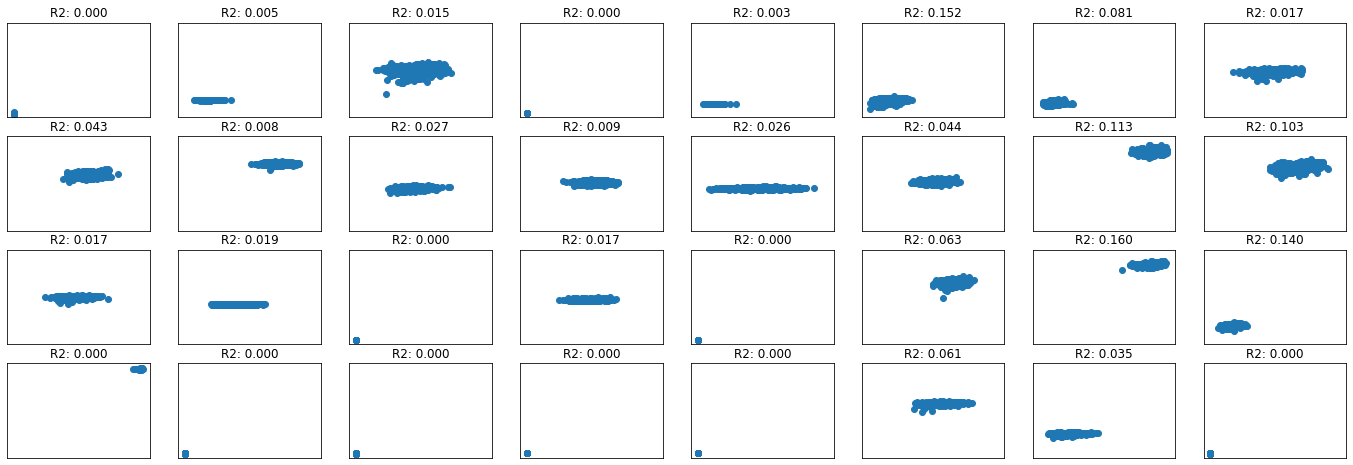}
    \caption{$\lambda=0.2$. The average $R^2$ statistic is 0.04, and 22 latent representations are learned.}
\end{subfigure}
\caption{Visualization of true latent representations (\textit{x-axis}) vs predicted latent representations (\textit{y-axis}) on the CelebA dataset}
\label{fig:covariance_plots}
\end{figure*}

Figure \ref{fig:covariance_plots} depicts each of the $32$ true latent representations vs their predictions for $1000$ samples of the CelebA dataset after three different NashAE networks have converged. When $\lambda=0$ (standard AE), all latent elements are employed towards the reconstruction objective, and the predictions exhibit a strong positive linear relationship with the true latent variables (average $R^2$ is $0.52$). When $\lambda=0.1$, only 28 latent representations are maintained, and the average $R^2$ statistic between true latents and their predictions becomes $0.09$. The 4 unused latent representations are each isolated in a dead zone of the sigmoid non-linearity, respectively. When $\lambda$ is increased to $0.2$, only $22$ latent representations are maintained and the average $R^2$ statistic decreases even further to $0.04$. Note also that the predictions become constant and are each equal to the expected value of their respective true latent feature. This is consistent with the conditional expectation of each variable being equal to its marginal expectation everywhere, and it indicates that no useful information is given to the predictors towards their regression task. 


\subsection*{Beamsynthesis Latent Space Visualization}

\begin{figure}[b]
\centering
\begin{subfigure}[t]{0.24\columnwidth}
    \centering
    \includegraphics[width=\textwidth]{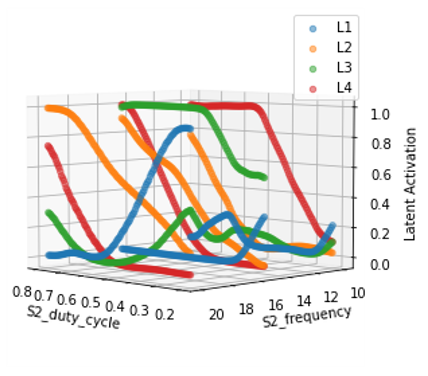}
    \caption*{NashAE $\lambda=0.0$}
\end{subfigure}
\begin{subfigure}[t]{0.24\columnwidth}
    \centering
    \includegraphics[width=\textwidth]{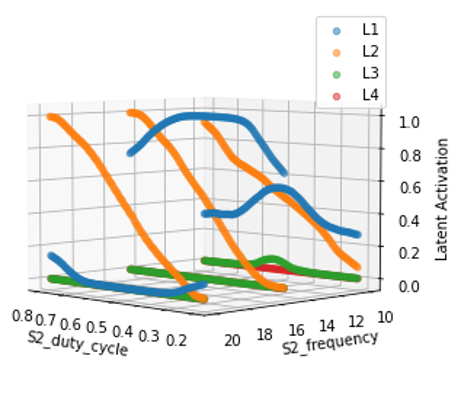}
    \caption*{NashAE $\lambda=0.2$}
\end{subfigure}
\begin{subfigure}[t]{0.24\columnwidth}
    \centering
    \includegraphics[width=\textwidth]{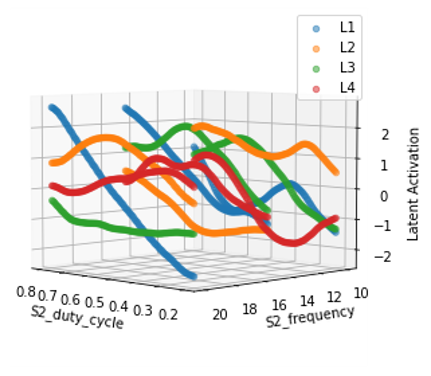}
    \caption*{$\beta$-VAE $\beta=1$}
\end{subfigure}
\begin{subfigure}[t]{0.24\columnwidth}
    \centering
    \includegraphics[width=\textwidth]{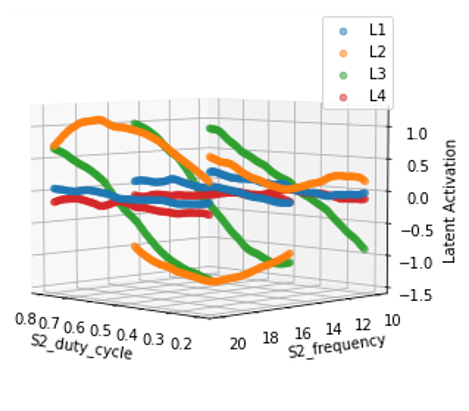}
    \caption*{$\beta$-VAE $\beta=100$}
\end{subfigure}
\caption{Visualizations of the learned latent space for the different algorithms on the Beamsynthesis dataset}
\label{fig:latent_vis}
\end{figure}

Figure \ref{fig:latent_vis} depicts the complete latent space generated by encoding all $360$ observations of the Beamsynthesis dataset for the different algorithms and their baselines with a starting latent size of $m=4$. 

A standard AE latent space (leftmost) employs all latent elements towards the reconstruction objective, and their relationship with the ground truth data generating factors, \textit{frequency} (categorical) and \textit{duty cycle} (continuous), is unclear. Similarly, when a standard VAE (center right) converges and the $\mu$ component of the latent space is plotted for all observations, all latent variables are employed towards the reconstruction objective, and no clear relationship can be established for the latent variables.

If an adversarial game is played with $\lambda=0.2$ (center left), the correct number of latent dimensions is extracted, and each nontrivial latent representation aligns with just one data generating factor. In this case, $L1$ level-encodes the \textit{frequency} categorical data generating factor, and $L2$ encodes the \textit{duty cycle} continuous data generating factor with a consistent interpretation. The unused neurons remain in a dead zone of the sigmoid non-linearity.

$\beta$-VAE with $\beta=100$ (rightmost) can disentangle the $360$ observations in a similar fashion. In this example, $L2$ level-encodes the \textit{frequency} categorical data generating factor, and $L3$ encodes the \textit{duty cycle} continuous data generating factor with a consistent interpretation. The unused neurons each have approximately $0$ variance in their $\mu$ component and an approximately constant value of $1$ in their learned variance component.

Although all algorithms are capable of extracting a disentangled representation of the ground truth data generating factors, there is a stark difference in the reliability of the methods in extracting the correct number of latent variables when the starting latent space size $m$ is changed. Reliability in this aspect is critical, as the dimensionality of the independent data generating factors is often an important unknown quantity to recover from new data. To determine this unknown dimensionality, one should start with a latent space size $m$ which is larger than the number of latent factors that should be extracted.

\begin{table}[hb!]
\begin{center}
\caption{Average absolute difference between the number of learned latent dimensions and the the number of ground truth factors for different starting latent space sizes $m$ on the Beamsynthesis dataset. Lower is better, and the lowest for each latent space size configuration ($m$) are in bold. The results are averaged over 8 trials}
\begin{tabular}{lcccc}
\hline
\noalign{\smallskip}
Method & $m=4$ & $m=8$ & $m=16$ & $m=32$ \\
\hline
\noalign{\smallskip}
NashAE $\lambda=0$ & 1.375 & 5.75 & 13.125 & 28.75 \\
NashAE $\lambda=0.2$ & \textbf{0} & \textbf{0} & \textbf{0.25} & 1 \\
NashAE $\lambda=0.3$ & \textbf{0} & \textbf{0} & 0.375 & \textbf{0.5} \\
\noalign{\smallskip}
\hline
\noalign{\smallskip}
$\beta$-VAE $\beta=1$ & 2 & 6 & 14 & 28.875 \\
$\beta$-VAE $\beta=50$ & 1.375 & 2 & 2.5 & 3.5 \\
$\beta$-VAE $\beta=100$ & \textbf{0} & 1 & 1.25 & 3.125 \\
$\beta$-VAE $\beta=125$ & 1.375 & 1.5 & 1.875 & 3.25 \\
\noalign{\smallskip}
\hline
\noalign{\smallskip}
$\beta$-TCVAE $\beta=1$ & 1.875 & 5.25 & 9.625 & 18.625 \\
$\beta$-TCVAE $\beta=50$ & 0.75 & 1.25 & 1 & \textbf{0.5}  \\
$\beta$-TCVAE $\beta=75$ & 0.375 & 0.5 & 1 & 0.875  \\
$\beta$-TCVAE $\beta=100$ & 0.75 & 0.625 & 1.125 & 0.625 \\
\noalign{\smallskip}
\hline
\noalign{\smallskip}
FactorVAE $\beta=50$ & 0.5 & 1.25 & 0.875 & 0.625 \\
FactorVAE $\beta=75$ & 0.5 & 1 & 1.25 & \textbf{0.5} \\
FactorVAE $\beta=100$ & 0.25 & 0.75 & 1 & 0.75 \\
FactorVAE $\beta=125$ & 0.875 & 0.75 & 0.625 & 0.75 \\
\noalign{\smallskip}
\hline
\noalign{\smallskip}

\end{tabular}
\centering
\label{table:lat_dim}
\end{center}
\end{table}

Table \ref{table:lat_dim} depicts the results of an experiment in which all hyperparameters are held constant except the starting latent size $m$ as each of the algorithms are trained to convergence on the Beamsynthesis dataset. Each entry in the table is the average absolute difference between the number of learned latent representations and the number of ground truth data generating factors (2 for Beamsynthesis), collected over 8 trials. Both NashAE $\lambda=0$ and $\beta$-(TC)VAE $\beta=1$ learn far too many latent variables, and $\beta$-VAE $\beta=125$ tends to learn too few latent variables when $m=4$ and $m=8$. NashAE $\lambda=0.2$ and NashAE $\lambda=0.3$ perform very well in comparison, keeping the average absolute difference less than or equal to one in all configurations of $m$. $\beta$-TCVAE and FactorVAE perform second-best overall, tending to learn too many latent variables. The results indicate that NashAE is the most consistent in recovering the correct number of data generating factors. See the supplementary material for a similar experiment with the dSprites dataset and details on how learned latent representations are counted.

\subsection*{$\beta$-VAE Metric on dSprites}

Table \ref{table:dsprites_metric} reports the disentanglement score of each algorithm averaged over 15 trials - please refer to Higgins \etal~\cite{higgins2016beta} for more details on the metric. In general, the standard AE (NashAE, $\lambda=0$) and standard VAE ($\beta$-VAE, $\beta=1$; $\beta$-TCVAE, $\beta=1$) performed the worst on the $\beta$-VAE disentanglement metric. As $\lambda$ and $\beta$ are increased, the disentanglement score of NashAE and $\beta$-VAE increases to over $96\%$. We do not observe the difference between NashAE and $\beta$-VAE in top performance on this metric to be significant, so both are in bold. In general, $\beta$-TCVAE performed slightly worse on this metric than $\beta$-VAE and NashAE, achieving just over $95\%$.  We observed that increasing $\lambda$ or $\beta$ beyond these values leads to poorer performance for all algorithms. All algorithms achieve higher disentanglement scores on some initializations than others, but no \textit{outliers} are removed from the reported scores (as is done in \cite{higgins2016beta}). Overall, the results indicate that NashAE scores at least as high as those of $\beta$-VAE and $\beta$-TCVAE algorithm on the $\beta$-VAE metric.

\begin{table}[ht!]
\centering
\caption{$\beta$-VAE Metric Scores on dSprites averaged over 15 trials. Higher is better, and the highest scores of all models and hyperparameter configurations are in bold. Optimal $\lambda$ values for disentanglement are different for this dataset because the dSprites image data is not normalized, following the precedent of previous works \cite{higgins2016beta}}
\begin{tabular}{lccc}
\noalign{\smallskip}
\hline
\noalign{\smallskip}
NashAE & $\lambda=0.0$ & $\lambda=0.001$ & $\lambda=0.002$ \\
   & 91.41\% & 92.58\% & \textbf{96.57\%} \\
\noalign{\smallskip}
\hline
\noalign{\smallskip}
$\beta$-VAE & $\beta=1$ & $\beta=4$ & $\beta=8$ \\
 & 84.63\% & 93.68\% & \textbf{96.21\% }\\
\noalign{\smallskip}
\hline
\noalign{\smallskip}
 $\beta$-TCVAE & $\beta=1$ & $\beta=2$ & $\beta=4$ \\
 & 84.64\% & 95.01\% & 93.95\% \\
\noalign{\smallskip}
\hline
\noalign{\smallskip}
\end{tabular}
\label{table:dsprites_metric}
\end{table}

\subsection*{Latent Traversals and TAD Metric on CelebA}

\begin{figure*}[hb!]
\begin{center}
\begin{subfigure}[b]{0.24\textwidth}
    \centering
    \includegraphics[width=\textwidth]{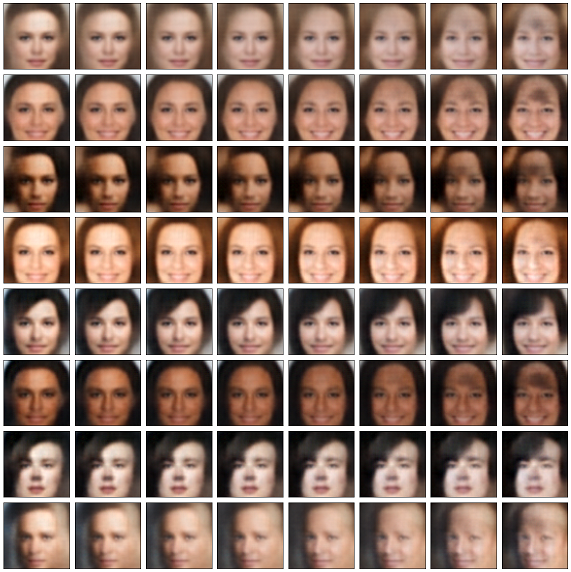}
    \caption*{\textbf{\textit{Bangs}} NashAE $\lambda=0.0$; Lat16 (0.788)}
\end{subfigure}
\begin{subfigure}[b]{0.24\textwidth}
    \centering
    \includegraphics[width=\textwidth]{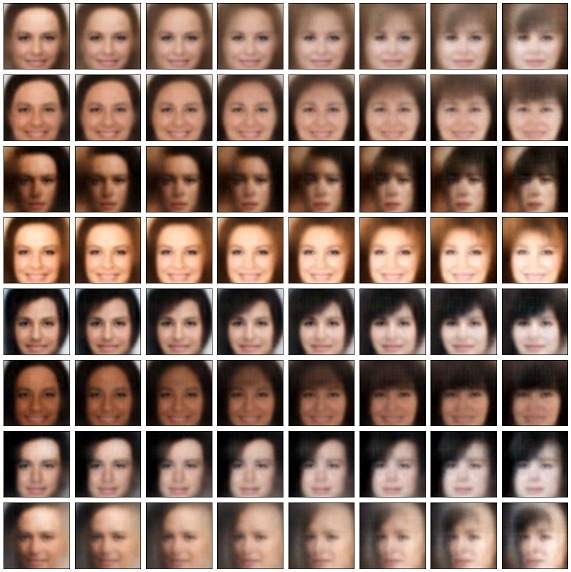}
    \caption*{\textbf{\textit{Bangs}} NashAE $\lambda=0.2$; Lat2 (0.831)}
\end{subfigure}
\begin{subfigure}[b]{0.24\textwidth}
    \centering
    \includegraphics[width=\textwidth]{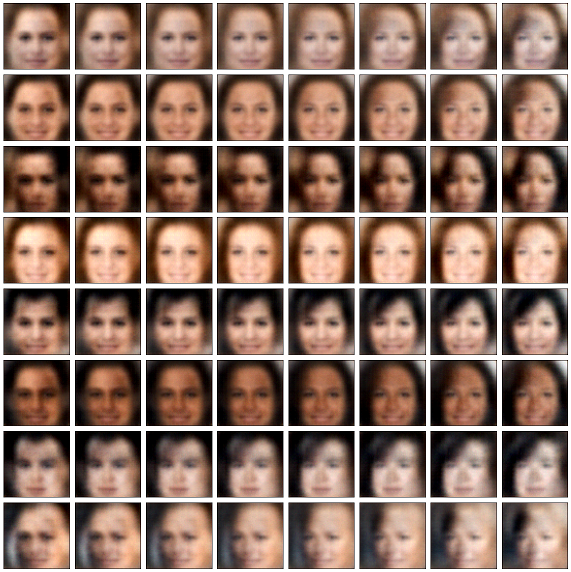}
    \caption*{\textbf{\textit{Bangs}} $\beta$-VAE $\beta=1$; Lat3 (0.701)}
\end{subfigure}
\begin{subfigure}[b]{0.24\textwidth}
    \centering
    \includegraphics[width=\textwidth]{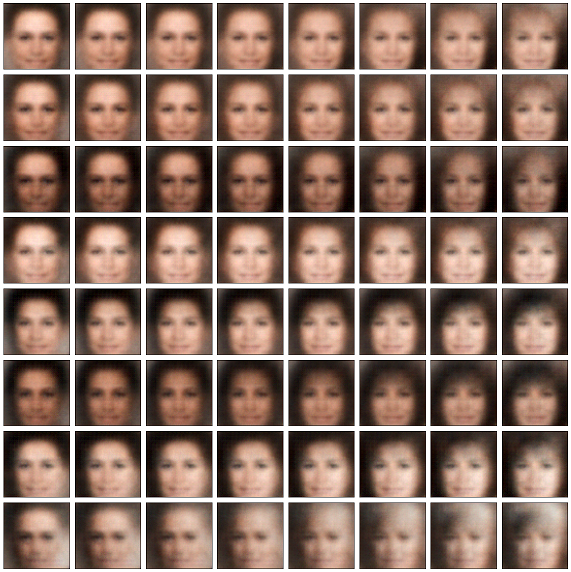}
    \caption*{\textbf{\textit{Bangs}} $\beta$-VAE $\beta=100$; Lat19 (0.724)}
\end{subfigure}
\end{center}
\begin{center}
\begin{subfigure}[b]{0.24\textwidth}
    \centering
    \includegraphics[width=\textwidth]{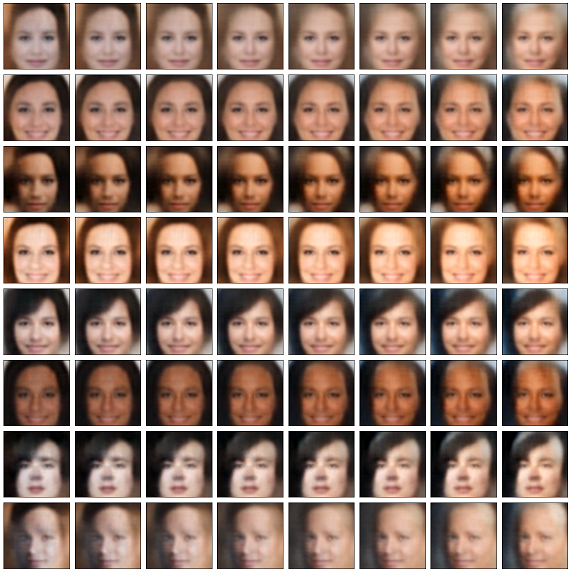}
    \caption*{\textbf{\textit{Blond}} NashAE $\lambda=0.0$; Lat28 (0.820)}
\end{subfigure}
\begin{subfigure}[b]{0.24\textwidth}
    \centering
    \includegraphics[width=\textwidth]{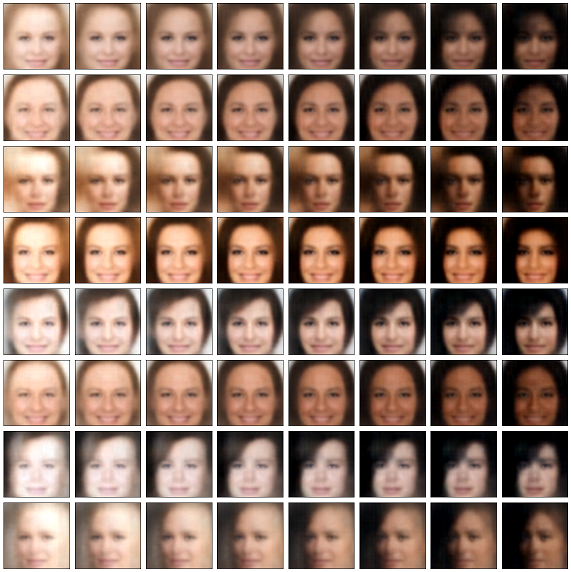}
    \caption*{\textbf{\textit{Blond}} NashAE $\lambda=0.2$; Lat12 (0.807)}
\end{subfigure}
\begin{subfigure}[b]{0.24\textwidth}
    \centering
    \includegraphics[width=\textwidth]{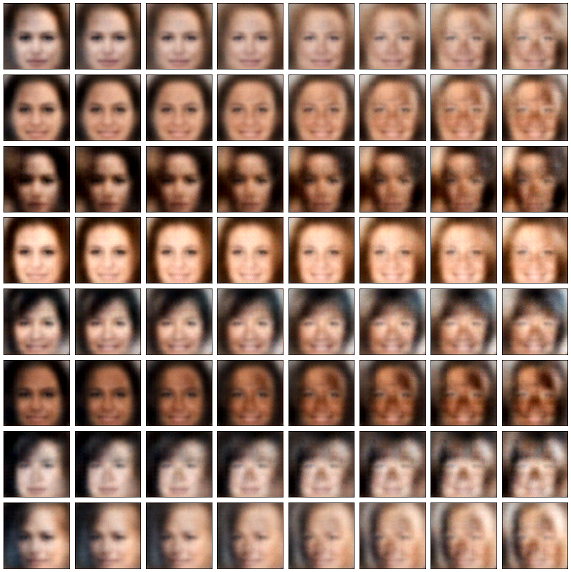}
    \caption*{\textbf{\textit{Blond}} $\beta$-VAE $\beta=1$; Lat18 (0.765)}
\end{subfigure}
\begin{subfigure}[b]{0.24\textwidth}
    \centering
    \includegraphics[width=\textwidth]{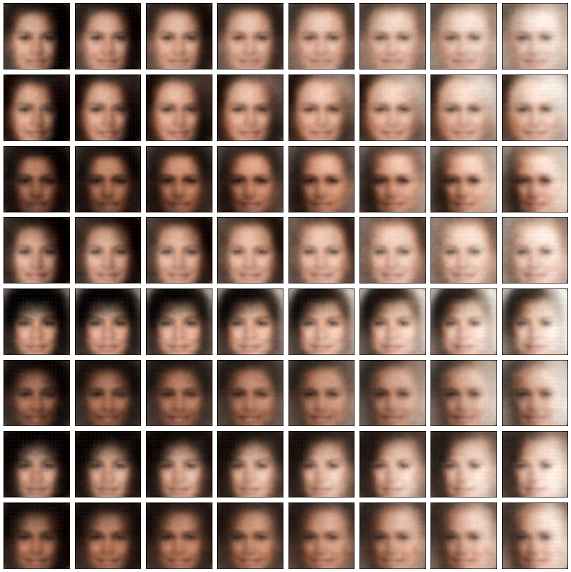}
    \caption*{\textbf{\textit{Blond}} $\beta$-VAE $\beta=100$; Lat11 (0.768)}
\end{subfigure}
\end{center}
\begin{center}
\begin{subfigure}[b]{0.24\textwidth}
    \centering
    \includegraphics[width=\textwidth]{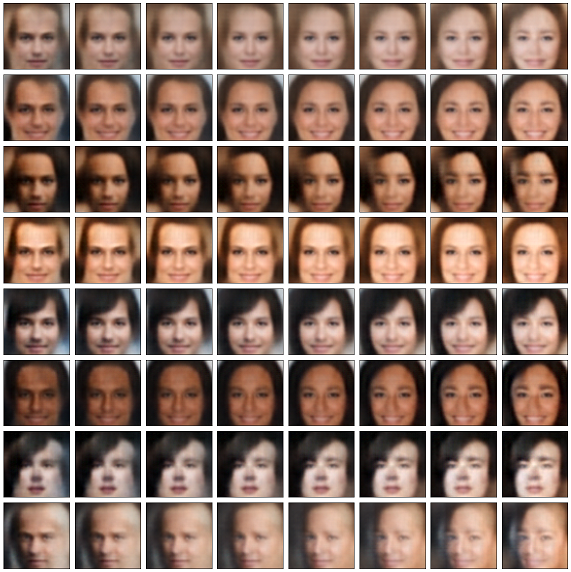}
    \caption*{\textbf{\textit{Male}} NashAE $\lambda=0.0$; Lat5 (0.664)}
\end{subfigure}
\begin{subfigure}[b]{0.24\textwidth}
    \centering
    \includegraphics[width=\textwidth]{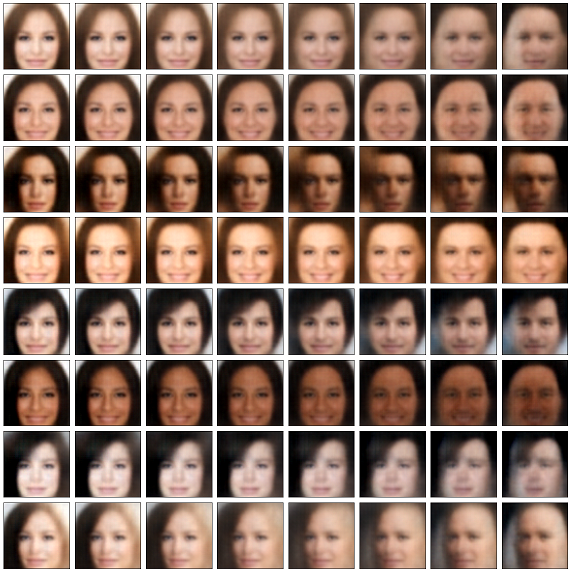}
    \caption*{\textbf{\textit{Male}} NashAE $\lambda=0.2$; Lat15 (0.697)}
\end{subfigure}
\begin{subfigure}[b]{0.24\textwidth}
    \centering
    \includegraphics[width=\textwidth]{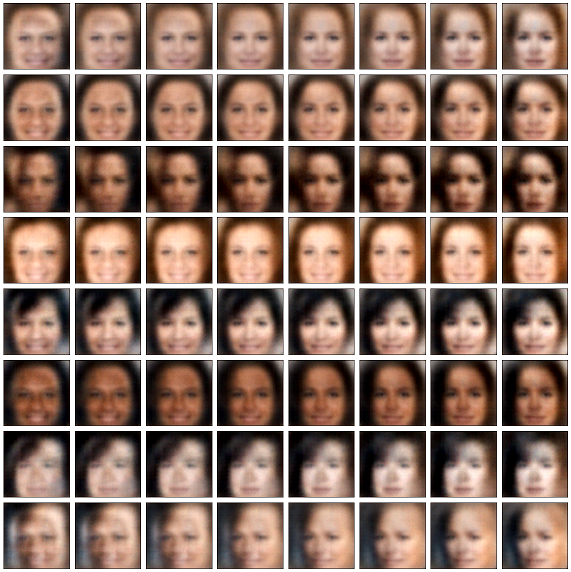}
    \caption*{\textbf{\textit{Male}} $\beta$-VAE $\beta=1$; Lat6 (0.645)}
\end{subfigure}
\begin{subfigure}[b]{0.24\textwidth}
    \centering
    \includegraphics[width=\textwidth]{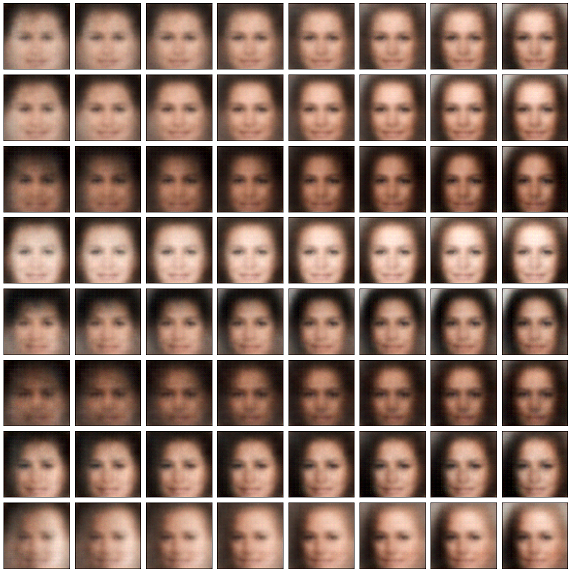}
    \caption*{\textbf{\textit{Male}} $\beta$-VAE $\beta=100$; Lat6 (0.632)}
\end{subfigure}
\end{center}
\caption{Traversals of latent features corresponding to the highest AUROC score for the \textbf{\textit{bangs}}, \textbf{\textit{blond}}, and \textbf{\textit{male}} attributes for the different disentanglement algorithms. Each latent representation with maximum AUROC score and its corresponding score are reported}
\label{fig:celeba_traversal}
\end{figure*}

We include traversals of latent representations that have the highest AUROC detector score for a small set of attributes on CelebA in Figure \ref{fig:celeba_traversal}. In each case, we start with a random image from the dataset and hold all latent representations constant except the one identified to have the highest AUROC score for the attribute of interest. We vary that representation evenly from its minimum to its maximum (as observed across 1000 random samples) and decode the resulting latent representation to generate the images reported in Figure \ref{fig:celeba_traversal}.

Note that in all cases, employing disentanglement methods (NashAE $\lambda > 0$ or $\beta$-VAE $\beta \gg 1$) leads to a visual traversal that intuitively matches the attribute that the latent representation is a good detector for. Furthermore, the visual changes are significant and obvious. Contrarily, when there is no effort to disentangle the representations ($\lambda=0$ or $\beta=1$), the relationship between the representation's high AUROC score and its traversal visualization become far less clear. In some cases, the traversal does not make meaningful change or even causes odd artifacts during decoding. We hypothesize that this is due to redundant information being shared between the latent features, and changing just one may have either no significant effect or the combination will be "out of distribution" to the decoder, leading to unnatural decoding artifacts. The idea that standard latents hold redundant information is supported by Figure \ref{fig:covariance_plots}, where the predictors establish a high average $R^2$ value on CelebA when $\lambda=0$. 
\begin{table}[b!]
\centering
\caption{TAD Scores on CelebA (averaged over 3 trials). Higher TAD scores are better, and the highest average score is in bold}
\begin{tabular}{lccc}
\hline\noalign{\smallskip}
Method & \ \ \ \ & TAD & \# Attributes \\
\noalign{\smallskip}
\hline
\noalign{\smallskip}
NashAE $\lambda=0$ & & 0.235 & 5.33 \\
NashAE $\lambda=0.1$ & & 0.362 & 4 \\
NashAE $\lambda=0.2$ & & \textbf{0.543} & 5 \\
NashAE $\lambda=0.8$ & & 0.474 & 5 \\
\noalign{\smallskip}
\hline
\noalign{\smallskip}
$\beta$-VAE $\beta=1$ & & 0.158 & 3.67 \\
$\beta$-VAE $\beta=50$ & & 0.287 & 2.67 \\
$\beta$-VAE $\beta=100$ & & 0.351 & 2.33 \\
$\beta$-VAE $\beta=250$ & & 0.307 & 2 \\
\noalign{\smallskip}
\hline
\noalign{\smallskip}
\end{tabular}
\quad
\begin{tabular}{lccc}
\noalign{\smallskip}\hline\noalign{\smallskip}
Method & \ \ \ \ & TAD & \# Attributes \\
\noalign{\smallskip}
\hline
\noalign{\smallskip}
$\beta$-TCVAE $\beta=1$ & & 0.165 & 3.33 \\
$\beta$-TCVAE $\beta=8$ & & 0.359 & 4 \\
$\beta$-TCVAE $\beta=15$ & & 0.446 & 4.33 \\
$\beta$-TCVAE $\beta=25$ & & 0.403 & 3.67 \\
$\beta$-TCVAE $\beta=50$ & & 0.362 & 3.67 \\
\noalign{\smallskip}\hline\noalign{\smallskip}
FactorVAE $\beta=1$ & & 0.188 & 3 \\
FactorVAE $\beta=8$ & & 0.208 & 2.33 \\
FactorVAE $\beta=15$ & & 0.285 & 3 \\
FactorVAE $\beta=50$ & & 0.276 & 3 \\
FactorVAE $\beta=75$ & & 0.148 & 1.33 \\
\noalign{\smallskip}\hline\noalign{\smallskip}
\end{tabular}
\centering
\label{table:total_auroc_diff}
\end{table}
We employ the TAD metric to quantify disentanglement on the CelebA dataset. Table \ref{table:total_auroc_diff} summarizes the TAD results and number of captured attributes for each of the algorithms averaged over three trials. An attribute is considered \textit{captured} if it has a corresponding latent representation with an AUROC score of at least $0.75$. The resulting scores indicate that the NashAE consistently achieves a higher TAD score, suggesting that its latent space captures more of the salient data characteristics (determined by the labelled attributes). Furthermore, NashAE achieves high scores over a broad range for $\lambda \in (0, 1)$. $\beta$-TCVAE performs second best, achieving a TAD score of $0.446$ when $\beta = 15$, yet it does not capture as many attributes as NashAE. In general, $\beta$-VAE and FactorVAE tend to capture fewer attributes and score lower TAD scores, suggesting that their latent spaces capture fewer of the salient data characteristics.

\section{Discussion}

We have shown with our quantitative experiments that NashAE can reliably extract disentangled representations. Furthermore, qualitative latent traversal inspection indicates that the latent variables of NashAE which are the best detectors for a given attribute indeed visually reflect independent traversals of the attribute. Hence, the adversarial covariance minimization objective presented in this work promotes learning of clarified, interpretable representations in neural networks. We believe that improvements in neural network interpretability can aid engineers in diagnosing and treating the current ailments of neural networks such as security vulnerability, lack of fairness, and out-of-distribution detection.

Future work will investigate more sophisticated latent distribution modeling and to make NashAE a generative model. This could further boost NashAE's disentanglement performance and provide deeper insight with information-theoretic approaches. It could be interesting to apply the adversarial covariance minimization objective to clarify the representations of DNNs for image classification.

\section{Conclusion}

We have presented NashAE, a new adversarial method to disentangle factors of variation which makes minimal assumptions on the number and form of factors to extract. We have shown that the method leads to a more statistically independent and disentangled AE latent space. Our quantitative experiments indicate that this flexible method is more reliable in retrieving the true number of data generating factors and has a higher capacity to align its latent representations with salient data characteristics than leading VAE-based algorithms.

\subsubsection*{Acknowledgements}

This research is supported, in part, by the U.S. Department of Energy, through the Office of Advanced Scientific Computing Research's “Data-Driven Decision Control for Complex Systems (DnC2S)” project. Additionally this research is sponsored by the Artificial Intelligence Initiative as part of the Laboratory Directed Research and Development Program of Oak Ridge National Laboratory (ORNL). This research used resources of the Experimental Computing Laboratory (ExCL) at ORNL.

This manuscript has been authored by UT-Battelle, LLC, under contract DE-AC05-00OR22725 with the US Department of Energy (DOE). The US government retains and the publisher, by accepting the article for publication, acknowledges that the US government retains a nonexclusive, paid-up, irrevocable, worldwide license to publish or reproduce the published form of this manuscript, or allow others to do so, for US government purposes. DOE will provide public access to these results of federally sponsored research in accordance with the DOE Public Access Plan (http://energy.gov/downloads/doe-public-access-plan).

This research is further supported by US Army Research W911NF2220025 and the National Science Foundation OIA-2040588.

\clearpage
%
%
\bibliographystyle{splncs04}
\bibliography{egbib}
\end{document}


\newcommand{\rulesep}{\unskip\ \vrule\ }

\newcommand{\etal}{et al.}

\pagestyle{headings}
\mainmatter
\def\ECCVSubNumber{6895}  

\subsection*{Limitations}

Like other unsupervised disentanglement methods such as $\beta$-VAE \cite{higgins2016beta} or $\beta$-TCVAE \cite{chen2018isolating}, some random initializations of the NashAE method resulted in better disentanglement scores than others, and training can fail in some very rare cases. This may be related to the findings of Locatello \etal~\cite{locatello2019challenging}, mentioned in the paper. Poor choices of hyperparameters such as a learning rate that is far too large can cause more frequent failures. No ``outliers'' were removed from the reported data in the paper, and the reported scores for all algorithms reflect honest averages.

Furthermore, each algorithm is trained with the same amount of data, and each autoencoder (AE) or variational AE (VAE) has roughly the same amount of parameters in each experiment (VAE has marginally more to implement the $\mu$, $\log(\sigma^2)$ parallel latent space). However, the NashAE method tends to collectively take more parameters and computations due to its use of the predictor ensemble. We observe that training for NashAE takes longer than VAE-based approaches due to the iterative predictor training. This process could be sped up by training the $m$ predictors in parallel.

Another limitation of NashAE is that it is not a generative model; it does not explicitly have a method to compute the likelihood of the data observations.

\subsection*{MinMax Formulation}

\textbf{Showing $\rvz' \rightarrow \E[\rvz]$.} If $\rvz$ and $\rvz'$ are independent, $\rvz' \rightarrow \E[\rvz]$. Consider $\rvz' = \E[\rvz] + \delta_{\rvz'}$, where $\E[\delta_{\rvz'}] = 0$. Plugging this into MISE($\rvz$, $\rvz'$), we get: $\frac{1}{2}\E[(\rvz - \E[\rvz] - \delta_{\rvz'})^2] = \frac{1}{2}\E[\rvz^2 + \E[\rvz]^2 - 2\rvz \E[\rvz] - 2\rvz \delta_{\rvz'} + 2\E[\rvz] \delta_{\rvz'} + \delta_{\rvz'}^2] = \frac{1}{2}(\E[\rvz^2] - \E[\rvz]^2 + \E[\delta_{\rvz'}^2]) = \frac{1}{2}(\sigma_\rvz^2 + \sigma_{\rvz'}^2)$. Since $\sigma_{\rvz'}^2 \geq 0$, this implies that SGD on $\rvz'$ reduces $\sigma_{\rvz'} \rightarrow 0$ to minimize the MISE objective. Hence, under these conditions, $\rvz' \rightarrow \E[\rvz]$.

\textbf{Partial Derivative of $\Cov(A, B)$.} Consider taking the partial derivative of the covariance between $K$ samples of two jointly distributed random variables $A$ and $B$ with respect to an observation $b_q$, where $1 \leq q \leq K$. Hence, $\frac{\partial}{\partial b_q}\Cov(A, B) = \frac{\partial}{\partial b_q} \frac{1}{K}\sum^K_{i=1}(a_i - \E[A])(b_i - \E[B])$. If $\E[A]$ and $\E[B]$ are computed as empirical averages across the $K$ samples, we have $\frac{\partial}{\partial b_q} \frac{1}{K}\sum^K_{i=1}(a_i - \E[A])(b_i - \E[B]) = \frac{1}{K}[(a_q - \E[A]) - \frac{1}{K}\sum^K_{i=1}(a_i - \E[A])] = \frac{1}{K}[(a_q - \E[A]) - \E[A] + \E[A]] = \frac{1}{K}(a_q - \E[A])$. Without loss of generality, this result can be applied to the statement of the next section.

\textbf{Equivalence between MSE and Covariance Objectives.} The fixed point of minimizing $\lambda\sum_{i=1}^m\Cov(\rvz'_i, \rvz_i)$ is equivalent to minimizing $-\frac{\lambda}{2}\E||\rvz' - \rvz||^2_2$ (i.e., maximize the MISE between $\rvz$ and $\rvz'$). This is because when a representation is disentangled, knowledge of other latent variables lends no useful information to the predictor. To minimize the MISE on its regression task, the uninformed $i$-th predictor will output $\E[\rvz_i]$ everywhere, making the gradient of $-\frac{\lambda}{2}\E||\rvz' - \rvz||^2_2$ equal to $\frac{\lambda}{K}(\rvz - \E[\rvz])$, which \textit{is equivalent to} the gradient term of minimizing $\lambda\sum_{i=1}^m\Cov(\rvz'_i, \rvz_i)$ with respect to the elements of an observation of $\rvz'$. In other words, $\frac{\partial}{\partial \rvz'} \lambda\sum_{i=1}^m\Cov(\rvz'_i, \rvz_i) = \frac{\partial}{\partial \rvz'} -\frac{\lambda}{2}\E||\rvz' - \rvz||^2_2$ if the elements of $\rvz$ are independent. Note that the AE uses the above gradient term along with $\frac{\partial \rvz'}{\partial \rvz}$ (evaluated at each sample in the batch) to adjust its representations $\rvz$ during the adversarial game (gradients are taken through the predictors only).

\subsection*{Experiment Setup}

All experiments were conducted using the PyTorch environment \cite{paszke2019pytorch}. Each algorithm is trained with the same amount of data - enough to ensure that both are converged for the given task. The learning rate is $0.001$ for the AE and the VAEs, and $0.01$ for the predictors in every experiment. We choose $k=5$ for the number of predictor update iterations per AE update. It is likely that $k$ can be reduced from $5$, but we did not explore other values of $k$ in this work. The details for each task are reported in table \ref{tab:experiment_setup}. SELU means self-normalizing activation function, from Klambauer \etal~\cite{klambauer2017self}. The signature for Conv is (\textit{width} x \textit{height})Conv(\textit{in\_channels}, \textit{out\_channels}, \textit{pad}, \textit{stride}). Recall that $m$ denotes the latent space size. The predictor architecture for every experiment is the following: Linear($m$, 40), SELU, Linear(40, 40), SELU, Linear(40, 1). The ReLU networks (for dSprites) are initialized via a Kaiming normal distribution \cite{he2015delving}. We employ the Adam optimizer \cite{kingma2014adam} with $\beta_1=0.9,\ \beta_2=0.999$ for all AEs, VAEs, and predictors.

\begin{table}[ht]
    \scriptsize
    \caption{Network architecture and hyperparameters for each experiment}
    \begin{tabularx}{\textwidth}{lcc}
        \hline
        Dataset & NashAE Architecture & VAE Architecture \\
        \hline
        \textbf{Beamsynthesis}  & Linear(1000, 200), SELU & Linear(1000, 200), SELU \\
        Mean/STD Data Norm.     & Linear(200, 80), SELU   & Linear(200, 80), SELU \\
                                & Linear(80, 40), SELU    & Linear(80, 40), SELU \\
        $BS \gets 100$          & Linear(40, $m$), Sigmoid & $2 \times$ Linear(40, $m$), (see VAE \cite{kingma2013auto}) \\
                                & Linear($m$, 40), SELU     & Linear($m$, 40), SELU \\
                                & Linear(40, 80), SELU & Linear(40, 80), SELU \\
                                & Linear(80, 200), SELU & Linear(80, 200), SELU \\
                                & Linear(200, 1000) & Linear(200, 1000), Gaussian \\
        \hline
        \textbf{dSprites} \cite{higgins2016beta} & Linear(4096, 1200), ReLU & Linear(4096, 1200), ReLU \\
        Flatten Tensor          & Linear(1200, 1200), ReLU & Linear(1200, 1200), ReLU \\
                                & Linear(1200, $m=10$), Sigmoid & $2 \times$ Linear(1200, $m=10$), (see VAE \cite{kingma2013auto}) \\
        $BS \gets 200$          & Linear($m=10$, 1200), ReLU   & Linear($m=10$, 1200), Tanh (see \cite{higgins2016beta}) \\
                                & Linear(1200, 1200), ReLU  & Linear(1200, 1200), Tanh \\
                                & Linear(1200, 4096)        & Linear(1200, 1200), Tanh \\
                                & - & Linear(1200, 4096), Bernoulli \\
        \hline
        \textbf{CelebA} \cite{liu2015faceattributes} & $4 \times 4$~Conv(3, 32, $1$, $2$), SELU & $4 \times 4$~Conv(3, 32, $1$, $2$), SELU \\
        Resize width to 96 & $4 \times 4$~Conv(32, 32, $1$, $2$), SELU & $4 \times 4$~Conv(32, 32, $1$, $2$), SELU \\
        Center Crop $64 \times 64$ & $4 \times 4$~Conv(32, 64, $1$, $2$), SELU & $4 \times 4$~Conv(32, 64, $1$, $2$), SELU \\
        Mean/STD Data Norm. & $4 \times 4$~Conv(64, 64, $1$, $2$), SELU & $4 \times 4$~Conv(64, 64, $1$, $2$), SELU \\
                & Linear(1024, 256), SELU & Linear(1024, 256), SELU \\
        $BS \gets 200$ & Linear(256, $m=32$), Sigmoid & $2 \times$ Linear(256, $m=32$), (see VAE \cite{kingma2013auto}) \\
                & Linear($m=32$, 256), SELU & Linear($m=32$, 256), SELU \\
                & Linear(256, 1024), SELU & Linear(256, 1024), SELU \\
                & $4 \times 4$~ConvT(64, 64, $1$, $2$), SELU & $4 \times 4$~ConvT(64, 64, $1$, $2$), SELU \\
                & $4 \times 4$~ConvT(64, 32, $1$, $2$), SELU & $4 \times 4$~ConvT(64, 32, $1$, $2$), SELU \\
                & $4 \times 4$~ConvT(32, 32, $1$, $2$), SELU & $4 \times 4$~ConvT(32, 32, $1$, $2$), SELU \\
                & $4 \times 4$~ConvT(32, 3, $1$, $2$) & $4 \times 4$~ConvT(32, 3, $1$, $2$), Gaussian \\
        \hline
    \end{tabularx}
    \label{tab:experiment_setup}
\end{table}

For FactorVAE training, the learning rate for the VAE and discriminator are both set to $10^{-4}$ for all experiments, following \cite{kim2018disentangling}. The architecture of the discriminator in all experiments is Linear($m$, $1024$), SELU, Linear($1024$, $1024$), SELU, Linear($1024$, $1024$), SELU, Linear($1024$, $1$), sigmoid. We employ the Adam optimizer \cite{kingma2014adam} with $\beta_1=0.5,\ \beta_2=0.9$, following \cite{kim2018disentangling}.

\subsection*{NashAE Algorithm}

See algorithm \ref{alg:training} for a detailed description of the NashAE training algorithm. 

\begin{algorithm*}[ht]
\centering
\begin{algorithmic}
\For{data minibatch $b$ in dataloader}
    \State $z \gets \phi(b)$ \Comment{$z$ is a minibatch of latent representations}
    \For{predictor $\rho_i$ in $\rho$}
        \State $\overline{\rvz_i} \gets \rvz \cdot (\underline{\textbf{1}} - \mathbb{I}_i); \forall \rvz \in z$ \Comment{$\mathbb{I}_i$ is the $i$-th column of the identity matrix}
        \For{number of predictor training iterations $k$}
        \State $\mathcal{L}_{\rho_i} \gets \frac{1}{2}\E_{\rvz \sim z}\big(\rho(\overline{\rvz_i})_i - \rvz_i\big)^2$ \Comment{reconstruction loss for the predictors}
        \State $\rho_i \gets \rho_i - \nu_\rho \nabla_{\rho_i}\mathcal{L}_{\rho_i}$ \Comment{SGD update for the $i$-th predictor's squared error using learning rate $\nu_\rho$}
        \EndFor
        \State $z_i' \gets \rho(\overline{\rvz_i})_i$ \Comment{prediction for the $i$-th latent variable, concatenates into $z'$} 
    \EndFor
    \State $b' \gets \psi(z)$ \Comment{$b'$ is the reconstructed data minibatch}
    \State $\mathcal{L}_R \gets \frac{1}{2n}\E_{\rvx \sim b}\big|\big|\rvx' - \rvx\big|\big|^2_2$ \Comment{reconstruction loss for the autoencoder}
    \State $\mathcal{L}_A \gets \sum_{i=1}^m\Cov(z'_i, z_i)$ \Comment{covariance of latent and prediction (across batch)}
    \State $\mathcal{L}_{R,A} \gets (1 - \lambda) \mathcal{L}_R + \lambda \mathcal{L}_A$ \Comment{combined loss for the AE}
    \State $(\phi,\psi) \gets (\phi,\psi) - \nu_{\phi,\psi}\nabla_{\phi,\psi} \mathcal{L}_{R,A}$ \Comment{SGD update for the AE, adversarial gradients are taken through $\rho$, learning rate $\nu_{\phi, \psi}$}
\EndFor
\end{algorithmic}
\caption{Training Algorithm for NashAE}\label{alg:training}
\end{algorithm*}

\subsection*{dSprites Dimensionality Experiment}

We present a dimensionality learning experiment similar to the one conducted with Beamsynthesis, but with the dSprites dataset in Table \ref{tab:dsprites_dim}. We include the VAE algorithms that were most competitive on Beamsynthesis. NashAE is still the most reliable in recovering the true number of data generating factors. Following the precedent of past works, we do not normalize the dSprites data, leading to ideal hyperparameter (e.g., $\lambda$ \& $\beta$) settings that differ from Beamsynthesis and CelebA.

\begin{table}[h!]
    \centering
    \caption{Average absolute difference between the number of factors learned by each algorithm and the true number of data generating factors (5) on the dSprites dataset}
    \begin{tabular}{l|c|c}
        \hline
        Algorithm & m=10 & m=20 \\
        \hline
        NashAE $\lambda=0.0$ & 5 & 15 \\
        NashAE $\lambda=0.004$ & 0.375 & 1 \\
        NashAE $\lambda=0.006$ & \textbf{0.25} & \textbf{0.375} \\
        NashAE $\lambda=0.008$ & 0.625 & 0.875 \\
        \hline
        FactorVAE $\beta=4$ &  0.875 & 1.875 \\
        FactorVAE $\beta=8$ &  0.5 & 0.875\\
        \hline
        $\beta$-TCVAE $\beta=1$ & 4.875 & 9.25 \\
        $\beta$-TCVAE $\beta=4$ & 0.5 & 1.125 \\
        $\beta$-TCVAE $\beta=8$ & 1.5 & 1.375 \\
        \hline
    \end{tabular}
    \label{tab:dsprites_dim}
\end{table}

\subsection*{Beamsynthesis Dataset Description}

This dataset is generated by a synthetic beamline model, which models the generation,
acceleration and tuning of ion beams in the LINAC (linear particle accelerator) portion of
high-energy particle accelerators. A more comprehensive description of the LINAC can be found in
references such as \cite{lee2018accelerator}. Here we only provide a brief description on its underlying mechanism.

The synthetic beamline model has three main components, which covers ion injection, the chopper and
the RF acceleration respectively. The ion injection component is based on the first-principle
physics model of ion sources, where particles are generated. The ion source cannot
be turned on instantaneously, therefore, there is a ramp-up phase of the ion beam, as depicted in the illustrative figures. This part of physics is modeled by six parameters. They are kept constant when the data were generated.

The chopper component models the beam chopper, which ``chops'' the ion beams into multiple
``pulses'' before they can be accelerated by the RF (radio frequency) cavities in the
LINAC. The speed the chopper rotates and relative opening determine the number of pulses as well as the the active region of each pulse. They are modeled by two parameters in the synthetic model: {\em frequency} and {\em duty\_cycle}. In the generated waveform,  parameter {\em frequency} represents a categorical latent space, while {\em duty\_cycle} represents a continuous latent space. These two parameters were permuted when the dataset were generated.

\begin{figure}[h]
    \centering
    \includegraphics[width=\textwidth]{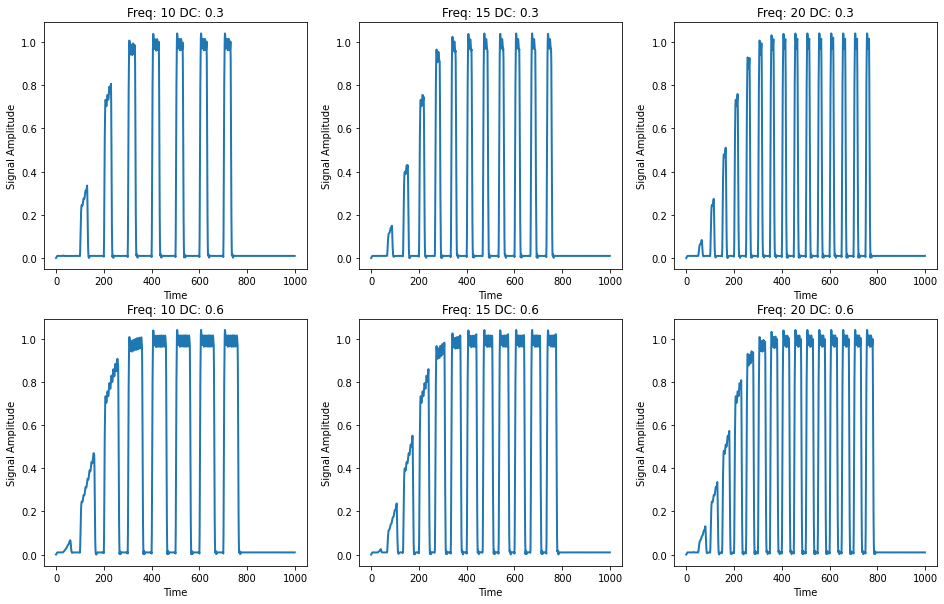}
    \caption{Depiction of varying \textit{frequency} (across a row) and \textit{duty\_cycle} (down a column) on the Beamsynthesis dataset. Freq: frequency, DC: duty\_cycle}
    \label{fig:beamsynthesis_vis}
\end{figure}

The RF acceleration component models the acceleration of the pulses by the RF cavities. The cavities not only
accelerate the particles to high speed, but also groups them into separate bundles. This component is modeled by five parameters, and were kept constant in the data generation. A set of beam waveform generated by the
synthetic model is shown in Figure. \ref{fig:beamsynthesis_vis}. A small amount of white noise has been added to reflect the stochastic nature of particle physics. Note that both x-axis (time) and y-axis (amplitude) have been normalized. Although only a handful of pulses are shown in the figure, hundreds of
thousands of pulses can be present in a real accelerator.

The Beamsyntheiss dataset has a few advantages. Since it is based on well-understood particle physics, one can adjust the parameters to reflect the underlying physics to control the latent space represented in the data. Moreover, since the synthetic model is quick to run, it is possible to generate large quantity of data for training and validation. We have found this dataset to be highly valuable in the development of the NashAE method. The dataset can be found at: \url{https://github.com/ericyeats/nashae-beamsynthesis}. 

\subsection*{Details on Learning Latent Feature Dimensionality (Beamsynthesis Dataset)}

A total of 8 independent trials are run for each algorithm and each set of hyperparameters. The absolute difference between the number of \textit{learned} latent features and the ground truth (2) is averaged between the 8 trials.

For this experiment, the presence of a \textit{learned latent feature} is determined in the following manner. For NashAE, the difference between the maximum and minimum latent scores for each latent feature is recorded. If the difference is larger than or equal to $0.2$, it is considered a \textit{learned latent feature}. Otherwise it is not counted. Likewise, for $\beta$-TCVAE and FactorVAE, a latent variable is considered \textit{learned} if the $\mu$ component of the VAE has a range greater than $2.0$. The $\beta$-VAE approach is inspired by how the authors count \textit{learned latent features} in the original work \cite{higgins2016beta}. If the average learned variance across the dataset is less than or equal to $0.8$, the latent is counted as a learned latent feature. Otherwise, if the average learned variance is greater than $0.8$, it is not counted. We also tried thresholding the maximum $\mu$ score minus the minimum $\mu$ score (similar to the criterion used for NashAE, $\beta$-TCVAE, and FactorVAE), but that resulted in consistently worse scores for $\beta$-VAE than the ones reported. We attribute this preferred method difference between $\beta$-VAE, $\beta$-TCVAE, and FactorVAE to the different weighting of the ELBO loss components during training.

\subsection*{$\beta$-VAE Metric Details}

Like in Higgins \etal~\cite{higgins2016beta}, we classify for \textit{x-position}, \textit{y-position}, \textit{scale}, and \textit{orientation} on the dSprites dataset. The linear classifier is initialized with a Kaiming Normal distribution \cite{he2015delving} and trained until convergence on batches of 100 examples each on a categorical crossentropy objective with a learning rate of $1.0$. For good measure, the learning rate is set to $0.05$ for the last 5\% of batches. The linear classifier converges in all tests. For training and testing, the constant data generating factor used to craft the batch is chosen with uniform probability (e.g.,~each factor gets a probability of 0.25 to be used to craft the batch). The linear classifier accuracy scores are collected over 1000 batches of size 100 each. Recall from Higgins \etal~\cite{higgins2016beta} that only the average absolute difference across each difference-batch is used to train and evaluate the classifier.

We also evaluated the algorithms with the $\beta$-VAE disentanglement metric on the Beamsynthesis dataset. Both methods achieved a 100\% score nearly 100\% of the time. This was also true for NashAE $\lambda=0$ and $\beta$-VAE $\beta=1$, and even when only one latent variable was learned by either method on this dataset that has two independent data generating factors. In the latter case, only one of the two factors was captured, but the linear classifier could tell which of the factors was held constant by the \textit{presence of} variation or \textit{lack of} variation in the single learned variable. In general, the high accuracy is due to the dataset being relatively simple.

\subsection*{Total AUROC Difference Details}

\textbf{Checking Reasonableness of TAD} We design a simple experiment for two synthetic latent variables $\rvz_\alpha$ and $\rvz_\beta$ and one bernoulli data generating factor $\rvc$ in which $p(\rvz_\alpha|c=1) = \mathcal{N}(\mu, 1)$ and $p(\rvz_\alpha|c=0) = \mathcal{N}(-\mu, 1)$. $p(\rvz_\beta)$ is distributed similarly to $p(\rvz_\alpha)$, but it is correlated to $c$ with strength $r$. When $\mu=1$ and $r=0.0$, TAD is 0.417. When $\mu=0.1$ and $r=0.0$, TAD is 0.07. When $\mu=1$ and $r=0.9$, TAD is 0.043. When $\mu=1$ and $r=-0.9$, TAD is 0.046. When $\mu=1$ and $r=0.0$, but there is a 1:50 true vs. false imbalance, TAD is 0.412. These experiments indicate that TAD increases for a confident, independent latent space, and it is not sensitive to class imbalance. We used 10k samples to compute TAD in each trial.

\textbf{TAD on CelebA} We measured TAD on the CelebA dataset \cite{liu2015faceattributes}. Since we are interested in measuring the TAD for only \textit{independent} attributes, we measure the proportion of entropy reduction of each attribute given knowledge of any other single attribute, and we disqualify any attribute with an entropy reduction greater than 0.2 from counting towards TAD. This disqualified attributes with high mutual information such as \textit{male}, \textit{wearing lipstick}, and \textit{wearing earings}. Furthermore, we are interested in measuring TAD for attributes which the algorithm is a good detector for. The algorithm is considered a good detector for an attribute if it has a maximum AUROC score of $0.75$. If the attribute does not make this cutoff, it is not considered ``learned'' by the algorithm and does not count. Lower thresholds for the maximum AUROC (e.g.~0.7) still resulted in the same general trends as those presented in the paper.

\textbf{How AUROC is Calculated} We test 100 evenly spaced detector threshold values from the minimum value to the maximum value for each latent on a large number of samples (e.g., 5000). We calculate the TPR and FPR for each of these 100 thresholds and both potential orientations (+ is increasing in latent values or + is decreasing in latent values) and take the maximum of the two AUROCs to ensure that the minimum AUROC score is 0.5.

\textbf{Comparison with other Work} The TAD metric is akin to the SAP metric proposed by Kumar \etal~\cite{kumar2017variational}, which uses difference in the factor classification accuracy of latent representations to measure the degree to which information of a single data generating factor is isolated in a single latent variable. However a key difference between SAP and TAD is that TAD is built off the AUROC score, which is threshold-independent and more informative for data with significant class imbalance (\textit{e.g.}, CelebA \cite{liu2015faceattributes}).

\subsection*{Reconstruction Quality of Different Methods}

We note that the reconstruction of NashAE tended to be better than that of the VAE-based methods, especially when $\beta >> 1$ in $\beta$-VAE. This is likely due to the information bottleneck component of the VAE loss, which both $\beta$-TCVAE and FactorVAE address by enhancing the total correlation (TC) component of the loss only.

\subsection*{Learned Latent Spaces on CelebA}

\begin{figure}
    \begin{center}
    \begin{subfigure}[b]{0.4\textwidth}
        \centering
        \includegraphics[width=\textwidth]{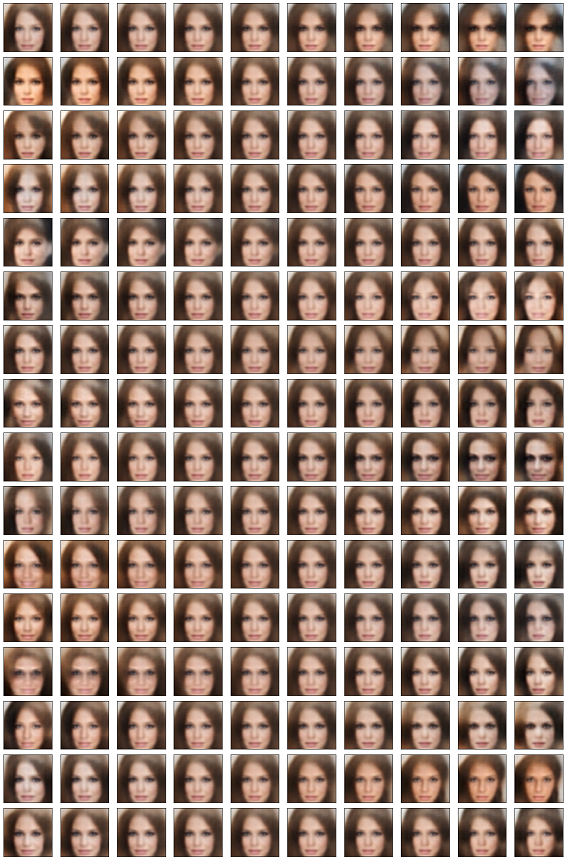}
    \end{subfigure}
    \begin{subfigure}[b]{0.4\textwidth}
        \centering
        \includegraphics[width=\textwidth]{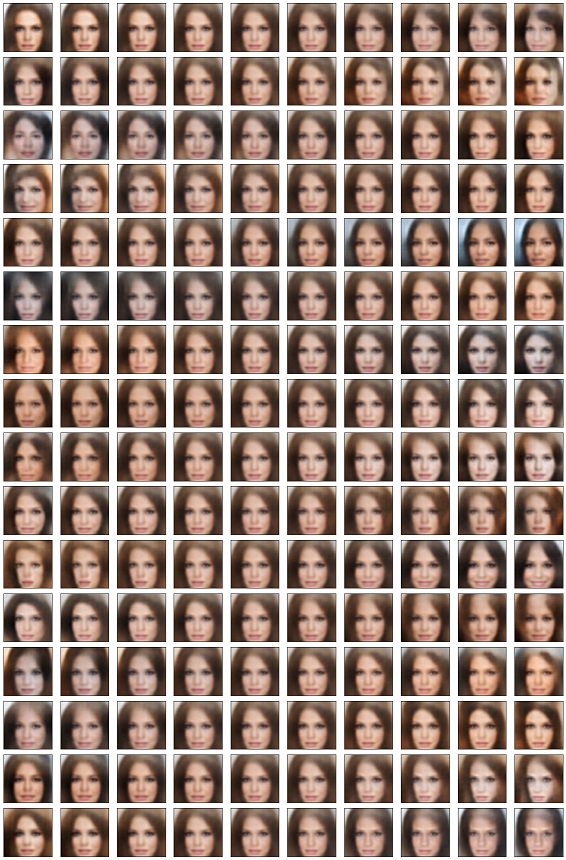}
    \end{subfigure}
    \end{center}
    \begin{center}
    \begin{subfigure}[b]{0.4\textwidth}
        \centering
        \includegraphics[width=\textwidth]{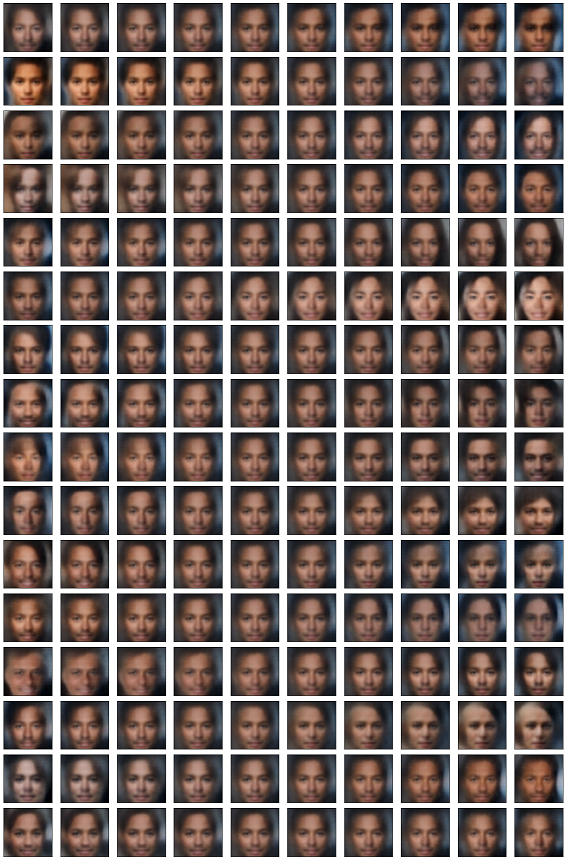}
    \end{subfigure}
    \begin{subfigure}[b]{0.4\textwidth}
        \centering
        \includegraphics[width=\textwidth]{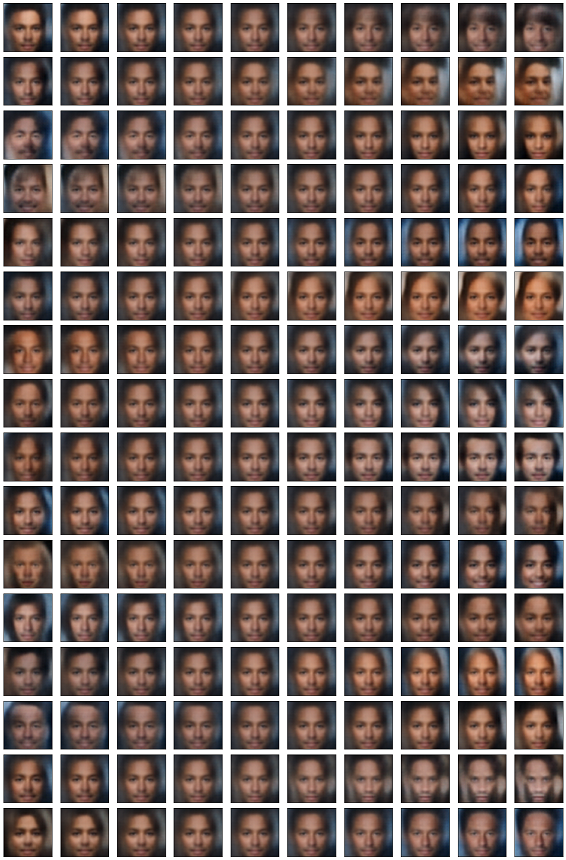}
    \end{subfigure}
    \end{center}
    \caption{Depiction of the latent space for NashAE $\lambda=0.0$ (Part A)}
    \label{fig:celeba_latents_0_A}
\end{figure}

\begin{figure}
    \begin{center}
    \begin{subfigure}[b]{0.4\textwidth}
        \centering
        \includegraphics[width=\textwidth]{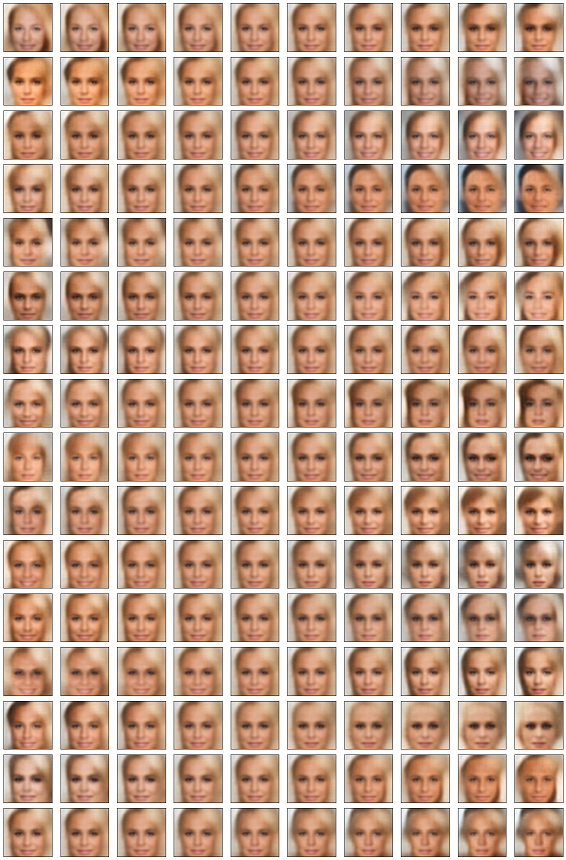}
    \end{subfigure}
    \begin{subfigure}[b]{0.4\textwidth}
        \centering
        \includegraphics[width=\textwidth]{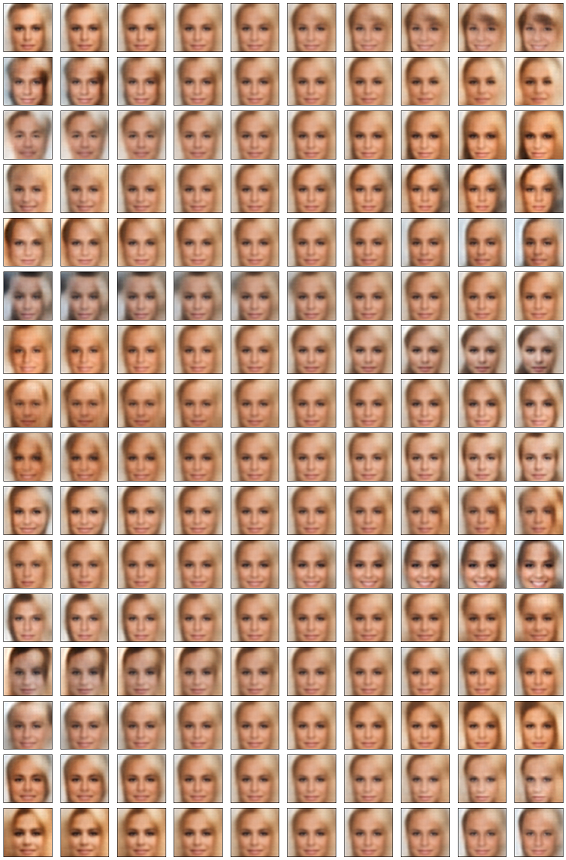}
    \end{subfigure}
    \end{center}
    \begin{center}
    \begin{subfigure}[b]{0.4\textwidth}
        \centering
        \includegraphics[width=\textwidth]{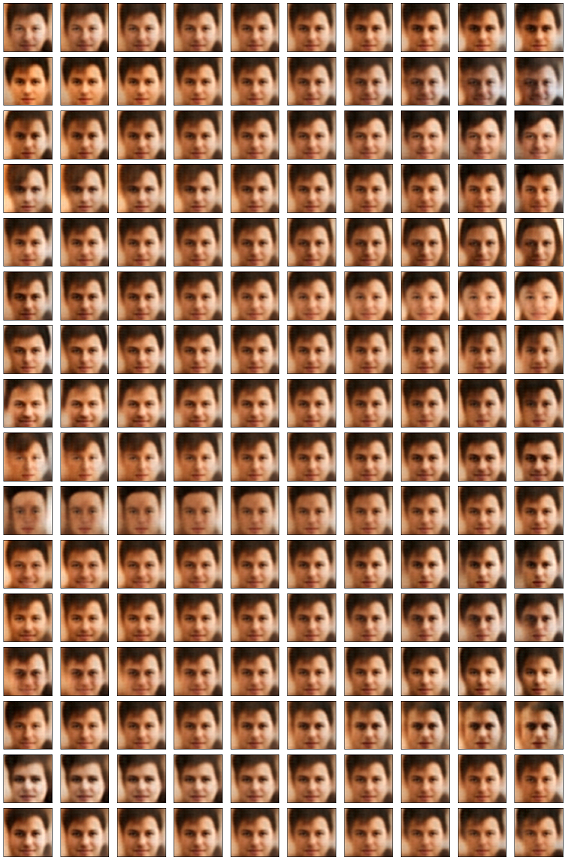}
    \end{subfigure}
    \begin{subfigure}[b]{0.4\textwidth}
        \centering
        \includegraphics[width=\textwidth]{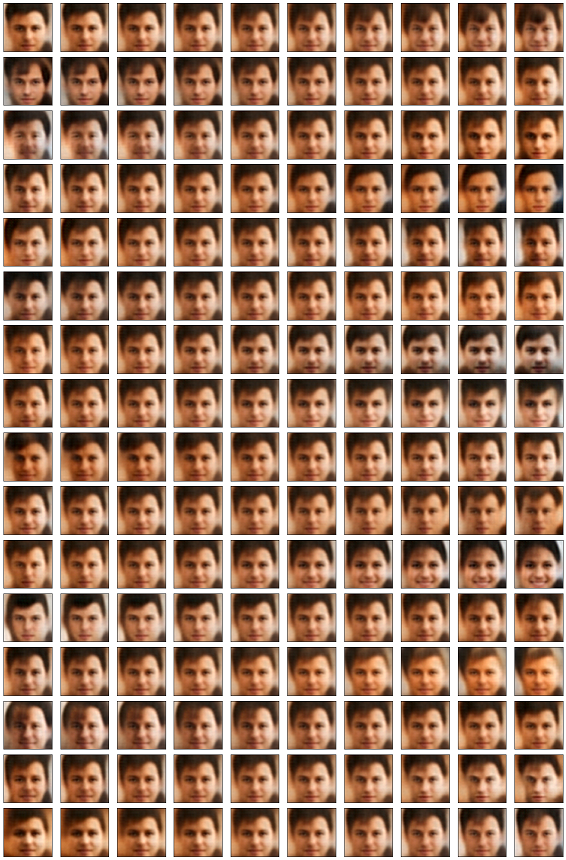}
    \end{subfigure}
    \end{center}
    \caption{Depiction of the latent space for NashAE $\lambda=0.0$ (Part B)}
    \label{fig:celeba_latents_0_B}
\end{figure}

\begin{figure}
    \begin{center}
    \begin{subfigure}[b]{0.4\textwidth}
        \centering
        \includegraphics[width=\textwidth]{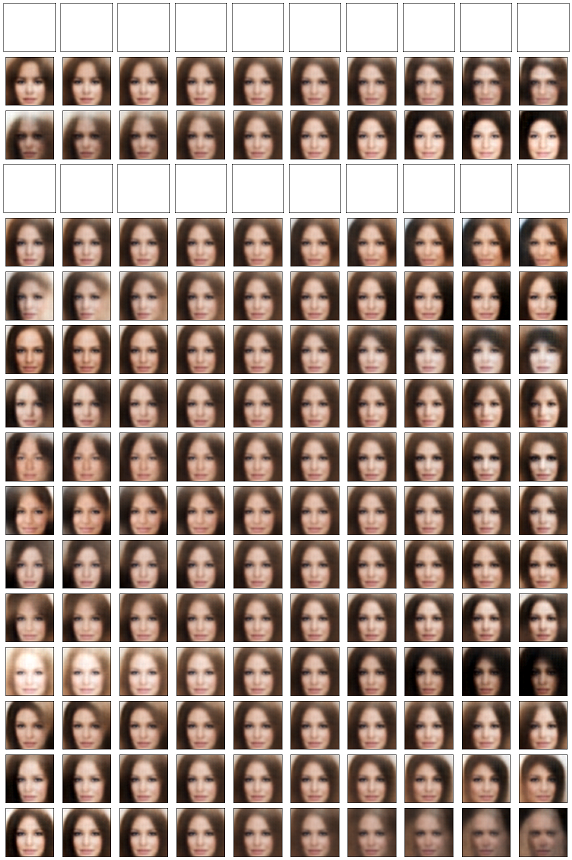}
    \end{subfigure}
    \begin{subfigure}[b]{0.4\textwidth}
        \centering
        \includegraphics[width=\textwidth]{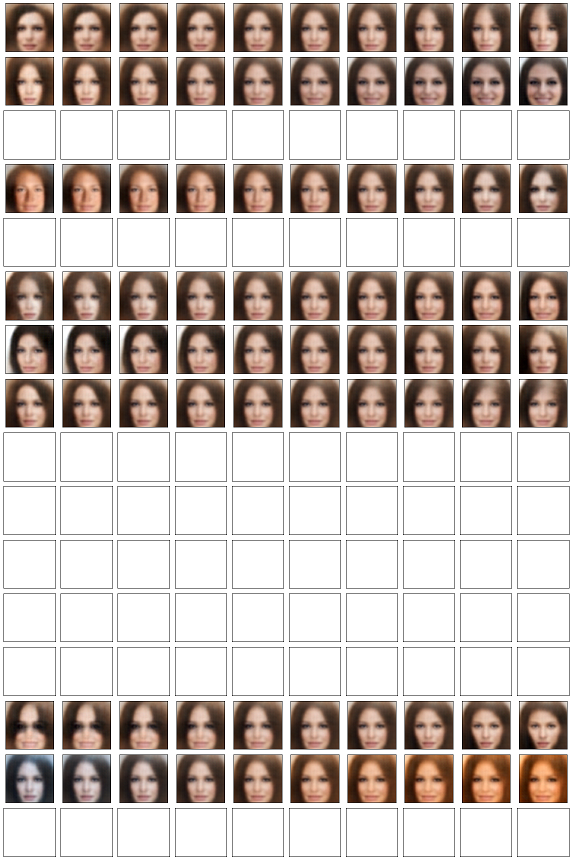}
    \end{subfigure}
    \end{center}
    \begin{center}
    \begin{subfigure}[b]{0.4\textwidth}
        \centering
        \includegraphics[width=\textwidth]{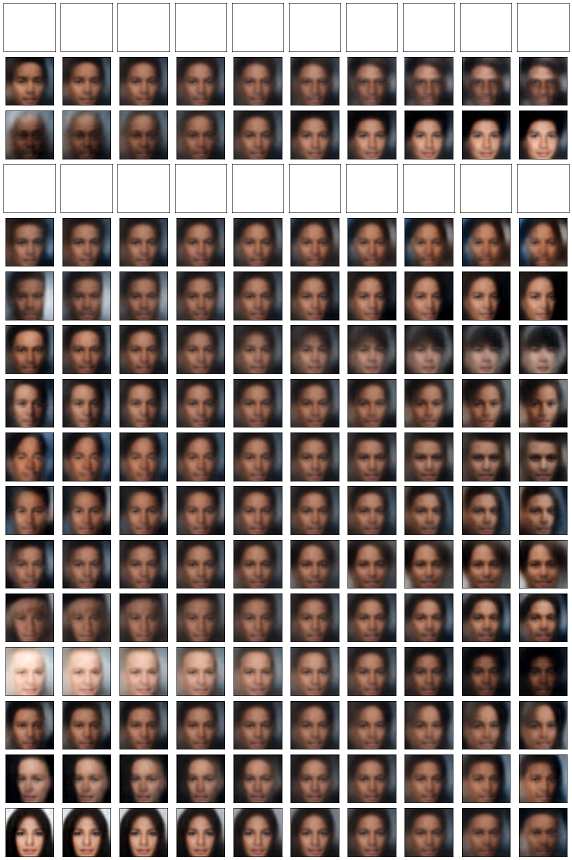}
    \end{subfigure}
    \begin{subfigure}[b]{0.4\textwidth}
        \centering
        \includegraphics[width=\textwidth]{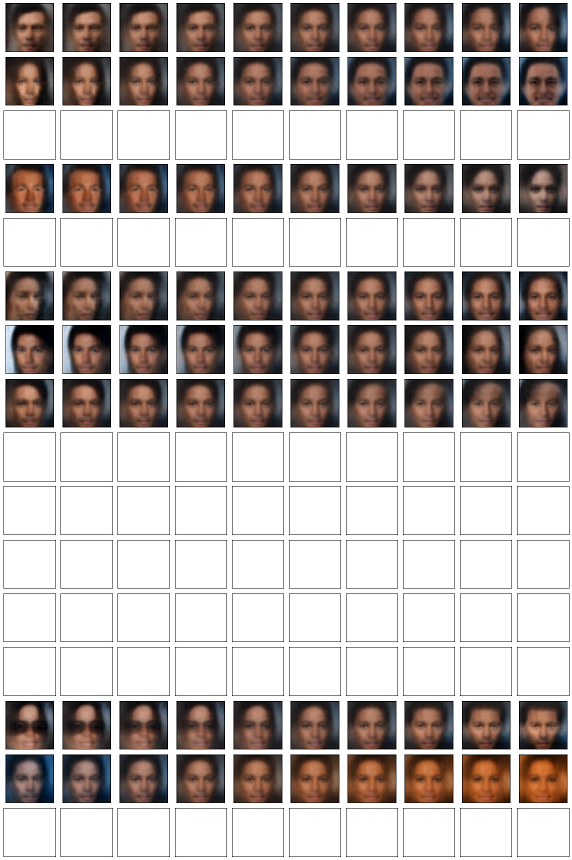}
    \end{subfigure}
    \end{center}
    \caption{Depiction of the latent space for NashAE $\lambda=0.2$ (Part A). Blank rows correspond to unused latent features}
    \label{fig:celeba_latents_2_A}
\end{figure}

\begin{figure}
    \begin{center}
    \begin{subfigure}[b]{0.4\textwidth}
        \centering
        \includegraphics[width=\textwidth]{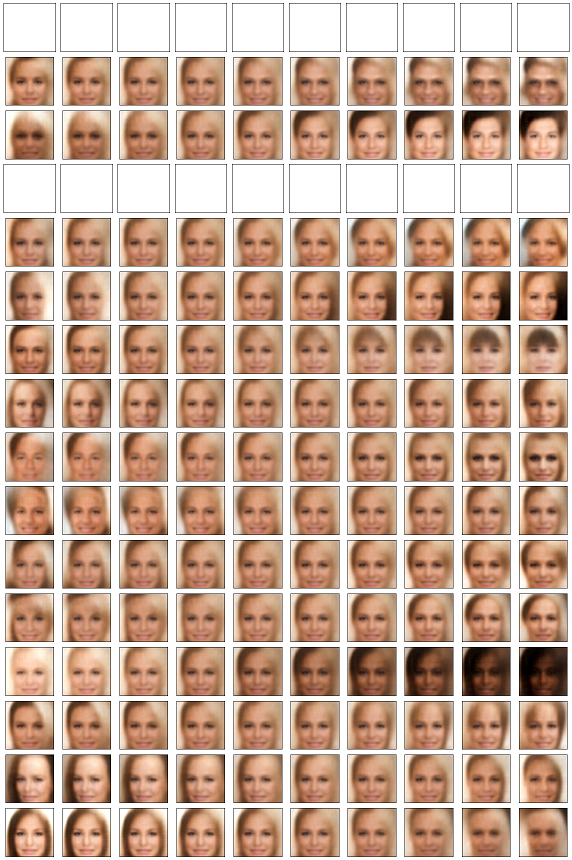}
    \end{subfigure}
    \begin{subfigure}[b]{0.4\textwidth}
        \centering
        \includegraphics[width=\textwidth]{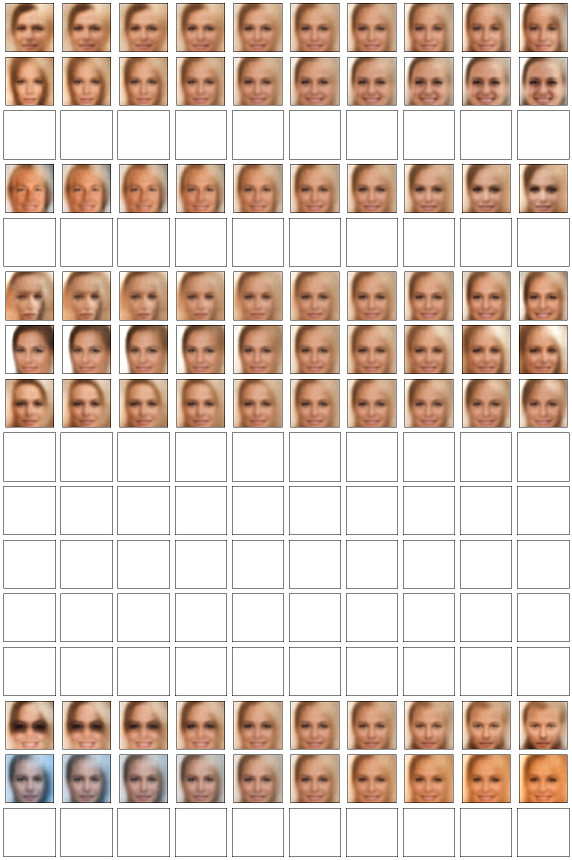}
    \end{subfigure}
    \end{center}
    \begin{center}
    \begin{subfigure}[b]{0.4\textwidth}
        \centering
        \includegraphics[width=\textwidth]{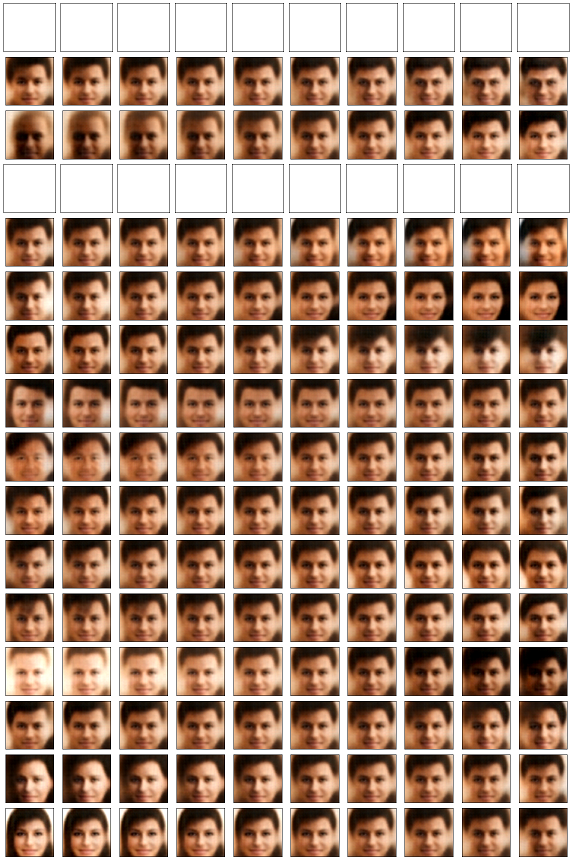}
    \end{subfigure}
    \begin{subfigure}[b]{0.4\textwidth}
        \centering
        \includegraphics[width=\textwidth]{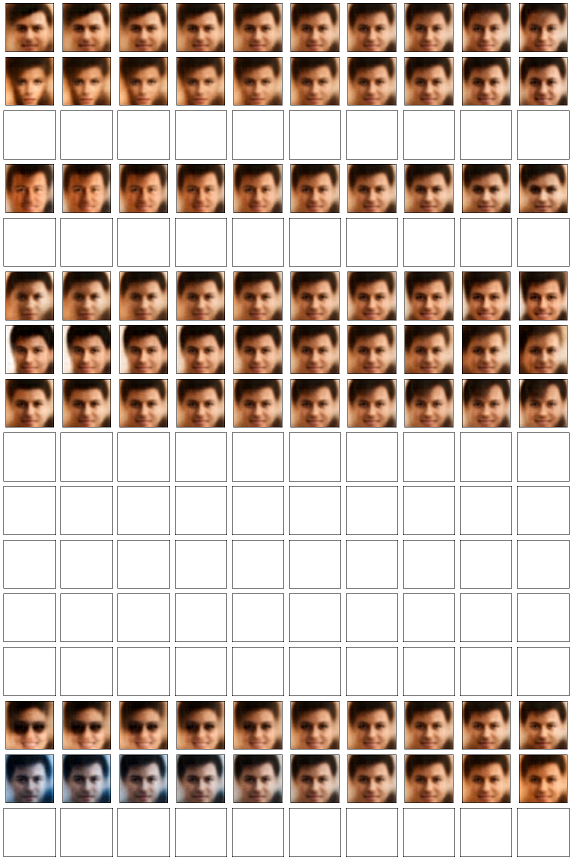}
    \end{subfigure}
    \end{center}
    \caption{Depiction of the latent space for NashAE $\lambda=0.2$ (Part B). Blank rows correspond to unused latent features}
    \label{fig:celeba_latents_2_B}
\end{figure}

We include full learned NashAE latent space traversals for the interested reader. See figures \ref{fig:celeba_latents_0_A}, \ref{fig:celeba_latents_0_B}, \ref{fig:celeba_latents_2_A}, and \ref{fig:celeba_latents_2_B}. Each row of 10 images corresponds to traversing a feature in the latent space while all others are held constant. Empty rows correspond to ``unused'' latent features.

%
%
\bibliographystyle{splncs04}
\bibliography{egbib}

